\documentclass[final]{opt2022} 

\usepackage{booktabs}
\usepackage{multirow}
\usepackage{hyperref}
\usepackage{boldline}
\usepackage[normalem]{ulem}
\useunder{\uline}{\ul}{}
\usepackage{amsmath}
\usepackage{todonotes}
\usepackage{comment}
\usepackage[nolist,nohyperlinks]{acronym}
\usepackage{cleveref}
\usepackage{dblfloatfix}
\usepackage{enumitem}
\setlist{nosep} 
\usepackage{pifont}
\usepackage{bm}

\usepackage{mathtools}

\title[Using Focal Loss to Fight Shallow Heuristics]{Using Focal Loss to Fight Shallow Heuristics:\\ An Empirical Analysis of Modulated Cross-Entropy in Natural Language Inference}

\optauthor{%
\Name{Frano Rajič} \Email{frano.rajic@epfl.ch}\\
\addr Swiss Federal Institute of Technology Lausanne (EPFL)
\AND
\Name{Ivan Stresec} \Email{ivan.stresec@gmail.com}\\
\addr Independent Researcher
\AND
\Name{Axel Marmet} \Email{axel.marmet@epfl.ch}\\
\addr Swiss Federal Institute of Technology Lausanne (EPFL)
\AND
\Name{Tim Poštuvan} \Email{tim.postuvan@epfl.ch}\\
\addr Swiss Federal Institute of Technology Lausanne (EPFL)
}

\begin{document}

\begin{acronym}[NLI] 
\acro{NLI}{natural language inference}
\acro{HANS}{Heuristic Analysis for NLI Systems}
\acro{MNLI}{MultiNLI}
\acro{SNLI}{the Stanford NLI corpus}
\end{acronym}

\maketitle

\begin{abstract}%

There is no such thing as a perfect dataset. In some datasets, deep neural networks discover underlying heuristics that allow them to take shortcuts in the learning process, resulting in poor generalization capability. Instead of using standard cross-entropy, we explore whether a modulated version of cross-entropy called focal loss can constrain the model so as not to use heuristics and improve generalization performance. Our experiments in natural language inference show that focal loss has a regularizing impact on the learning process, increasing accuracy on out-of-distribution data, but slightly decreasing performance on in-distribution data. Despite the improved out-of-distribution performance, we demonstrate the shortcomings of focal loss and its inferiority in comparison to the performance of methods such as unbiased focal loss and self-debiasing ensembles.




\ifanonsubmission
\else
  Code available at \texttt{\href{https://github.com/m43/focal-loss-against-heuristics}{github.com/m43/focal-loss-against-heuristics}}.
\fi

\end{abstract}


\section{Introduction}


The quality of a neural network is often directly related to the quality of the dataset it was trained on \cite{focal-loss-in-cv-2017, agrawal2016vqa, dodge2016understanding, imagenetC}. The optimization process inherently exploits any shortcuts available, which can result in undesirable heuristics being learned. In \cite{xiao2021noise}, for example, the authors have found that the models rely on various heuristics when classifying images from ImageNet. In the field of natural language processing, there exist many other examples where models learn only shallow heuristics instead of obtaining a more profound understanding of the task \cite{naik2018stress, sanchez2018behavior, min-etal-2019-compositional, schwartz-etal-2017-effect, gururangan2018annotation}. Such models are flawed as they cannot generalize well in out-of-distribution data, implying that they are likely to perform poorly in real-world scenarios.

If simple heuristics can be exploited in the training dataset, the loss will provide little to no incentive for the model to generalize to unseen data, data for which the heuristics will not work, resulting in negative learning outcomes. We hypothesize that focal loss \cite{focal-loss-in-cv-2017} can be used to alleviate this generalization issue by giving more weight to wrongly classified samples. Assuming that the samples which do not adhere to heuristics have low true prediction probabilities, and assuming that they will not get memorized by an expressive model, the focal loss could potentially amplify the otherwise insufficient supervision signal that comes from underrepresented samples.


We use the \ac{NLI} \cite{bowman2015large} task to investigate our hypothesis and our experiments show that the focal loss outperforms the status-quo cross-entropy loss on an out-of-distribution dataset. However, the accuracy on hard samples of domain test sets is generally decreased, suggesting that the focal loss alone fails to improve performance on more difficult examples. Furthermore, our analysis of the probability distribution  of the networks' predictions shows that the focal loss produces a more uncertain network. We contribute this fact to the focal loss's property of higher penalization of difficult samples at the price of lowering the reward of a network's prediction certainty. Despite the improved out-of-distribution performance, the shortcomings of focal loss demonstrate its inferiority in comparison to the performance of methods such as unbiased focal loss \cite{bias-mitigation-2020} and self-debiasing ensembles \cite{utama2020towards,sanh2021learning,clark2020learning}, suggesting that the latter approaches should be preferred in most similar settings.


\section{Related Work}

\subsection{Tackling Heuristics in NLI}


Several works have proposed effective debiasing methods that work well on natural language understanding tasks including the \ac{NLI} task, but not strictly from an optimization perspective.

McCoy et al. \cite{right-wrong-2019} create a dataset that identifies whether a model learns one of the targeted heuristics. By injecting samples from the designed dataset into the training data, models are discouraged from learning a chosen heuristic. While the idea works well, its major drawback is the requirement of manually detecting heuristics and creating a new dataset accordingly.

By contrast, Mahabadi et al. \cite{bias-mitigation-2020} propose using an auxiliary model to discover biases of the dataset and to automatically adapt the relative importance of examples for the base model. The auxiliary model is combined with the base models using two approaches, both of which result in improved performance on adversarial test datasets. The first approach utilized is using a product of experts.
The second approach is using unbiased focal loss that the authors introduce, and that downweights the loss for the samples that the auxiliary model has learned. Both approaches require \textit{a priori} knowledge on designing an auxiliary model that can capture the heuristics of interest (\textit{e.g.}, in the \ac{NLI} task, such a model might take only the hypothesis as input, without having the premise as input). The approach we explore (\textit{i.e.}, focal loss) is completely model and dataset agnostic, and can be implemented by merely substituting the loss function. It can be seen as an ablation study of the debiased focal loss, since we isolate focal loss and examine it on its own, without the debiasing factors coming from a hand-designed auxiliary model.

Subsequent work \cite{utama2020towards,clark2020learning,sanh2021learning,liu2021just} moves away from the limitation of knowing the dataset biases \textit{a priori}. 
For example, \cite{utama2020towards} introduces a self-debiasing framework that complements existing methods. A shallow model is trained as an auxiliary model and then used to downweight the potentially biased samples when training the base model. Such an auxiliary model does not require knowledge of the biases beforehand in designing its architecture. On a similar note, Liu et al. \cite{liu2021just} manage to improve worst-group error. Instead of using shallow models, the authors train one model for just several epochs and then train a second model that upweights the training samples that the first model misclassified. 


\subsection{Heuristics as Class Imbalance} 
The existence of heuristics that the majority of samples of a dataset adhere to can be seen as an analogy to class imbalance, in which the minority class corresponds to samples not adhering to heuristics. Lin et al. \cite{focal-loss-in-cv-2017} introduce focal loss and demonstrate that it allows the training of dense object detectors with higher accuracy, despite the presence of an abundance of easy samples that outnumber the hard samples by about $10^4$ times. We try using the same approach on \ac{NLI}, but with a slightly different goal and without using $\alpha$-balancing.




\subsection{Learning from Underspecified Data}

One of the datasets we work with has been shown to be underspecified \cite{d2020underspecification} since the training distribution does not contain enough clues about performing \ac{NLI} correctly. Recent work \cite{lee2022diversify} tackles the problem of underspecified data by proposing a simple two-stage framework that first learns a diverse set of hypotheses for solving a task, and then, based on additional supervision, selects one hypothesis that promises better generalization performance. The authors demonstrate the robustness of the selected hypothesis for tasks in image classification and natural language processing. 

Similar to this, \cite{pagliardini2022agree} introduces an algorithm that enforces models in an ensemble to disagree on out-of-distribution data, but agree on the training data. Based on experimental results, the method mitigates shortcut-learning, enhances uncertainty and out-of-distribution detection, and improves transferability.



\section{Methodology}

\subsection{Focal Loss}
\begin{figure*}[t]
    \centering
    \includegraphics[width=\textwidth]{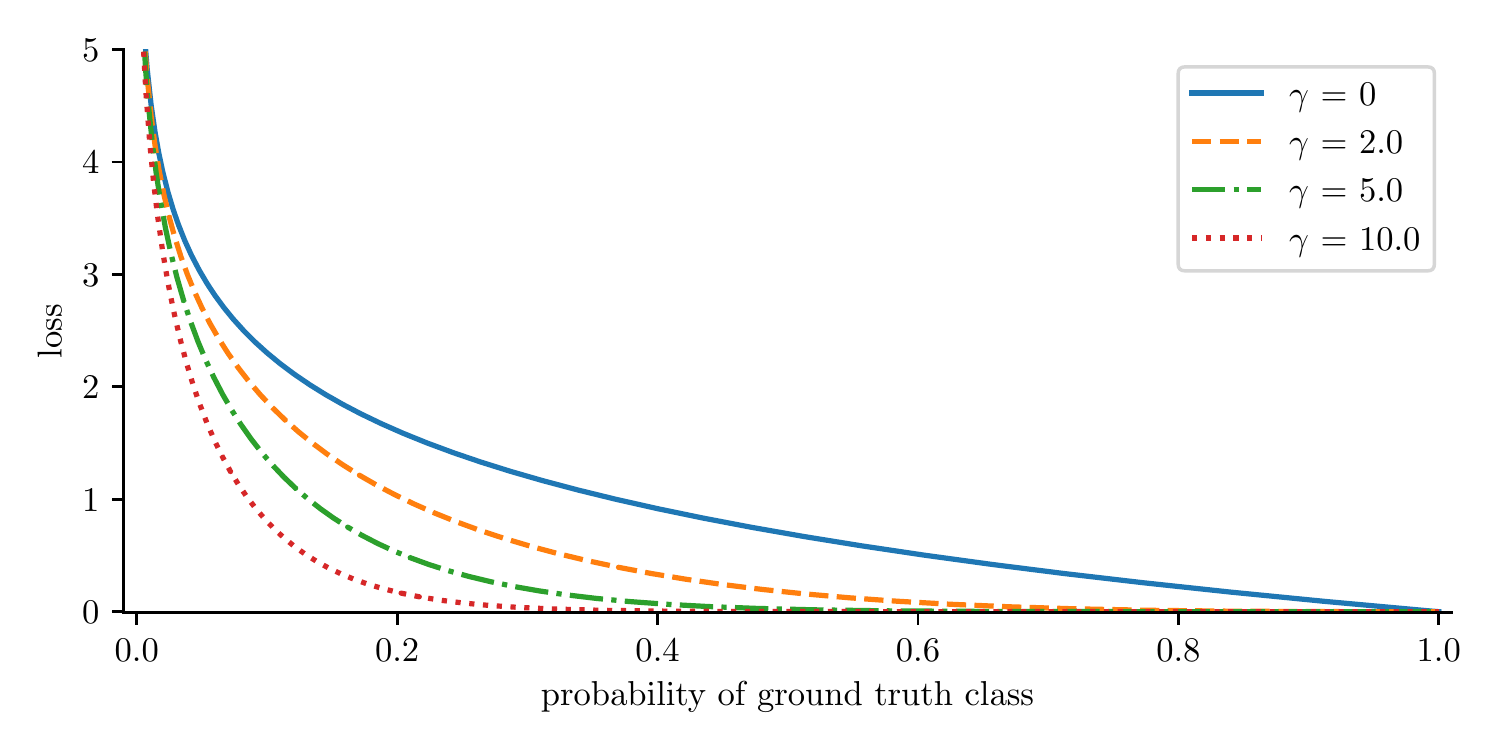}
    \caption{A graph showing the effect of different $\gamma$ parameters for the focal loss function, with $\gamma  = 0$ being equal to regular cross-entropy.}
    \label{fig:focal_loss_gamma}
\end{figure*}

Focal loss \cite{focal-loss-in-cv-2017} was originally designed to alleviate the class imbalance problem during training by applying a modulating term to the cross-entropy loss, as a way of putting a higher emphasis on misclassified samples. In turn, however, focal loss also down-weights the contribution of simpler and well-classified samples during training. The focal loss of a single sample is calculated in the following manner:
\begin{equation}
    FL(p_t) = -(1 - p_t)^{\gamma}\log(p_t),
\end{equation}
where $p_t$ is the model's output probability for the ground truth label $y$ and $\gamma$ is a parameter that alters the modulating term $(1 - p_t)$. With higher values of $\gamma$, focal loss places less importance on samples that are already correctly and confidently classified and thus allows the model to focus on harder samples that could potentially contradict simple heuristics. Note that when $\gamma = 0$, the loss function is equivalent to the cross-entropy loss. The impact of a higher-valued $\gamma$ parameter can best be seen in Figure \ref{fig:focal_loss_gamma}.

Loss functions like the Huber Loss \cite{esl} are designed to mitigate the influence of outliers in the dataset, making the training more robust to noise. Focal Loss does the contrary by emphasizing misclassified samples during training, which could be counterproductive for noisy datasets. 



\subsection{Heuristic Analysis for NLI Systems (HANS) Dataset}
\label{hansabs}

In \ac{NLI}, the objective is to determine whether a premise sentence entails a hypothesis sentence (e.g., the statement \textit{"The doctor was visited by him."} entails the statement \textit{"He visited the doctor."}). The \ac{HANS} dataset \cite{right-wrong-2019} provides a benchmark to test whether a model solves the \ac{NLI} task by having learned one of the three heuristics shown in Table \ref{tab:hans_heuristics}.

The \ac{HANS} dataset is primarily a diagnostic tool since the models that learn the analysed heuristics do not generalize well and will produce a $50\%$ accuracy (accuracy of 0\% for the adversary \textbf{non-entailed (NE)} samples and 100\% for the \textbf{entailed (E)} ones). The dataset can further be used to stop the model from capitalizing on these heuristics by adding a small proportion of \ac{HANS} samples into the training dataset.

In our experiments, we utilize \ac{HANS} for two purposes: (1) as a test dataset; and (2) to investigate the impact of using focal loss when \ac{HANS} samples are added to the training dataset, as compared to models trained with cross-entropy loss.

\begin{table}[t]
\centering
\caption{Definitions of heuristics that the \ac{HANS} dataset is analysing and examples of (premise $\rightarrow$ hypothesis) pairs demonstrating when the heuristic would fail to recognize non-entailment. The examples are randomly taken samples from the HANS dataset.}
\begin{tabular}{p{0.11\linewidth} p{0.345\linewidth} p{0.46\linewidth}}
    \toprule
    Heuristic &  Definition &  Example \\
    \toprule
    Lexical overlap \newline (L)
    & Assume that a premise entails all hypotheses that contain only words from the premise.
    & \textbf{The secretary} and \textbf{the judge supported} the lawyer.  \newline $\xrightarrow[\text{WRONG}]{}$ The secretary supported the judge. \\
    \bottomrule
    Subsequence \newline (S) & Assume that a premise entails all of its contiguous subsequences. & Before \textbf{the doctor paid the author} arrived. \newline $\xrightarrow[\text{WRONG}]{}$ The doctor paid the author. \\
    \bottomrule
    Constituent \newline (C) & Assume that a premise entails all complete subtrees in its parse tree. & Hopefully \textbf{the senators mentioned the artists}. \newline $\xrightarrow[\text{WRONG}]{}$ The senators mentioned the artists. \\
    \bottomrule
\end{tabular}
\label{tab:hans_heuristics}
\end{table}

\section{Experiments}

\subsection{Models and Datasets}
We explore the performance of using focal loss on two models: BERT \cite{bert} with a classification head \cite{bert, right-wrong-2019} and InferSent \cite{infersent, bias-mitigation-2020}.
Furthermore, we train the two models on two datasets: (1) \ac{SNLI} \cite{snli}, a standard \ac{NLI} dataset of approximately $570,000$ annotated sentence pairs and (2) \ac{MNLI} \cite{mnli}, a multi-genre \ac{NLI} dataset of approximately $433,000$ annotated sentence pairs modelled after \ac{SNLI}. 

The use of \ac{MNLI} dataset is especially interesting to our hypothesis since the dataset is almost completely oblivious to counterexamples to the syntactic heuristics. We verified that it contains only about 250 counterexamples to the heuristics, similar to what was observed by McCoy et al. \cite{right-wrong-2019}. The dataset bias in both \ac{SNLI} and \ac{MNLI} is inherited from a crowdsourcing process in which crowd workers tend to write the hypothesis sentences by adding simple modifications to the given premise sentence \cite{he2019unlearn}. For example, to forge a hypothesis that contradicts the premise, crowd workers save time by adding words such as "no", "none", "never", or "nothing" to the premise \cite{gururangan2018annotation}. Simple heuristics that look only at the hypothesis sentence can therefore achieve high performance on these datasets, but will fail on more realistic data, especially the \ac{HANS} dataset, due to different data distributions. Accordingly, we use the HANS validation subset as a test set for measuring the effectiveness of focal loss in reducing the learning of shallow heuristics. Furthermore, we measure the performance on hard examples from the test sets of \ac{MNLI} and \ac{SNLI} separately, in order to see focal loss's impact when dealing with more difficult samples. The hardness labels are based on the ability of a shallow model to correctly classify hypotheses without being given the premise \cite{joulin2017bag, gururangan2018annotation, bias-mitigation-2020}.

\subsection{Implementation}
We implemented the BERT model using the HuggingFace library and adapt InferSent's code from \cite{bias-mitigation-2020}. Due to the cost of training, we train models using hyperparameters from previous work - they are listed in Appendix \ref{sec:appendix1}. However, we do not fix the number of epochs like in \cite{bias-mitigation-2020}, but instead use early stopping based on the loss of the validation data, retaining model parameters with the lowest validation loss value. We have observed that using early stopping can cause subpar performance on test sets in comparison to that in the mentioned literature, but decided to use it nevertheless as we could not justify fixing the number of epochs otherwise.

The \ac{MNLI} dataset has two validation sets: the matched and the mismatched set. The mismatched set contains samples that do not closely resemble those in the training set. We use the mismatched validation set for early stopping models trained on the MNLI dataset, hoping that the distribution shift will provide a more relevant measure of the performance for the test sets and real-world application. The matched validation subset is, however, used for testing and refer to it as MNLI Test in this paper. Note that we do not use the original MNLI matched and mismatched test subsets, since they are withheld from the public and are only accessible through online benchmarks.

\subsection{Experimental Design}
We perform four sets of experiments, one for each of the models and the datasets, repeating each experiment $5$ times to obtain a measure of the second momentum. The results of these can be seen in Table \ref{tab:exp1}. Additionally, for the BERT model we perform experiments in which we add $100$ and $1000$ samples from the \ac{HANS} training set. In this way, we examine focal loss when simulating a reduced underspecification of the datasets with regard to the HANS test set. In other words: when the distribution shift between the training and test sets is less significant. These results can be seen in Table \ref{tab:exp2}.

\begin{table}[t]
\centering
\begin{minipage}{.53\linewidth}
\caption{Test set acccuracy of the BERT and InferSent models trained on the \ac{MNLI} and \ac{SNLI} datasets. For \ac{MNLI}, we have used the mismatched validation subset for early stopping, and the matched validation subset for testing. \ac{SNLI} has both the validation and test set publicly available, so we use them directly. Hardness of samples is defined following \cite{gururangan2018annotation}. We report the average accuracy across $5$ runs alongside the corresponding standard deviation. More detailed results for HANS are in the Appendix (Table \ref{tab:exp1hans}).}
\label{tab:exp1}
\centering
\resizebox{1\textwidth}{!}{
\begin{tabular}{@{}cccccc@{}}
\toprule
         && \multicolumn{2}{c}{\textbf{MNLI}} & \multicolumn{2}{c}{\textbf{SNLI}} \\  
         \cmidrule(r){3-4} \cmidrule(l){5-6} 
                 && \textbf{BERT} & \textbf{InferSent} &\textbf{BERT} & \textbf{InferSent} \\
         \toprule 
         \multicolumn{6}{c}{\textbf{Test set results}}\\
         \toprule
 \parbox[t]{2mm}{\multirow{6}{*}{\rotatebox[origin=c]{90}{\textbf{Gamma}}}}
     & 0.0 & $ \textbf{83.82}\pm0.3 $ & $ \textbf{69.45}\pm0.4 $ & $ 90.31\pm0.2 $ & $ \textbf{83.83}\pm0.3 $ \\
     & 0.5 & $ 83.81\pm0.3 $ & $ 69.12\pm0.3 $ & $ \textbf{90.37}\pm0.1 $ & $ 83.50\pm0.2 $ \\
     & 1.0 & $ 83.71\pm0.4 $ & $ 69.00\pm0.4 $ & $ 90.14\pm0.3 $ & $ 83.43\pm0.3 $ \\
     & 2.0 & $ 83.38\pm0.3 $ & $ 68.42\pm0.3 $ & $ 89.90\pm0.2 $ & $ 82.97\pm0.4 $ \\
     & 5.0 & $ 82.65\pm0.3 $ & $ 67.50\pm0.4 $ & $ 89.07\pm0.3 $ & $ 81.43\pm0.2 $ \\
     & 10.0 & $ 81.84\pm0.3 $ & $ 64.49\pm0.4 $ & $ 88.59\pm0.4 $ & $ 79.31\pm0.3 $ \\
         \toprule 
         \multicolumn{6}{c}{\textbf{HARD Test subset results}}\\
         \toprule
 \parbox[t]{2mm}{\multirow{6}{*}{\rotatebox[origin=c]{90}{\textbf{Gamma}}}}
     & 0.0   & $ \textbf{76.08}\pm0.4 $ & $ \textbf{55.91}\pm0.5 $ & $ 79.74\pm0.5 $ & $ \textbf{84.75}\pm0.3 $ \\
     & 0.5 & $ 76.06\pm0.4 $ & $ 55.63\pm0.4 $ & $ \textbf{79.77}\pm0.4 $ & $ 84.36\pm0.4 $ \\
     & 1.0   & $ 75.80\pm0.7 $ & $ 55.08\pm0.5 $ & $ 79.36\pm0.5 $ & $ 84.40\pm0.4 $ \\
     & 2.0   & $ 75.32\pm0.4 $ & $ 54.64\pm0.3 $ & $ 78.95\pm0.4 $ & $ 83.95\pm0.5 $ \\
     & 5.0   & $ 74.35\pm0.5 $ & $ 53.49\pm0.5 $ & $ 77.50\pm0.2 $ & $ 82.61\pm0.4 $ \\
     & 10.0  & $ 73.50\pm0.5 $ & $ 50.47\pm0.5 $ & $ 76.53\pm0.4 $ & $ 80.50\pm0.4 $ \\
         \toprule 
         \multicolumn{6}{c}{\textbf{HANS}}\\
         \toprule
 \parbox[t]{2mm}{\multirow{6}{*}{\rotatebox[origin=c]{90}{\textbf{Gamma}}}}
     & 0.0   & $ 52.20\pm1.0 $ & $ 50.68\pm0.3 $ & $ 58.08\pm2.3 $ & $ 50.61\pm0.4 $ \\
     & 0.5 & $ 52.82\pm1.3 $ & $ 50.94\pm0.4 $ & $ 57.17\pm1.5 $ & $ 51.14\pm0.6 $ \\
     & 1.0   & $ 53.61\pm2.0 $ & $ 51.02\pm0.2 $ & $ 56.99\pm2.8 $ & $ 51.35\pm0.3 $ \\
     & 2.0   & $ 54.91\pm3.1 $ & $ 51.48\pm0.7 $ & $ 59.49\pm1.5 $ & $ 51.58\pm0.3 $ \\
     & 5.0   & $ \textbf{59.54}\pm3.2 $ & $ \textbf{53.15}\pm0.7 $ & $ \textbf{63.33}\pm2.8 $ & $ \textbf{52.47}\pm0.7 $ \\
     & 10.0  & $ 56.22\pm5.8 $ & $ 49.85\pm0.2 $ & $ 58.71\pm6.0 $ & $ 49.99\pm0.3 $ \\ \bottomrule
\end{tabular}%
}
\end{minipage}%
\begin{minipage}{.03\linewidth}\enspace\end{minipage}
\begin{minipage}{.43\linewidth}
\caption{Test set accuracy of the BERT model trained on \ac{MNLI} with random \ac{HANS} training samples added to the training set. The test set refers to the MNLI matched validation set, and MNLI mismatched validation set was used for early stopping. Hardness of samples is defined following \cite{gururangan2018annotation}. We report the average accuracy across $5$ runs alongside the corresponding standard deviation. More detailed results for HANS are in the Appendix (Table \ref{tab:exp2hans}).}
\label{tab:exp2}
\centering
\resizebox{1\linewidth}{!}{%
\begin{tabular}{@{}ccccc@{}}
\toprule
&& \multicolumn{3}{c}{\textbf{\# HANS samples added to train}} \\  
&& $\textbf{0}$ & $\textbf{100}$ & $\textbf{1000}$ \\

         \toprule 
         \multicolumn{5}{c}{\textbf{Test set results}}\\
         \toprule
 \parbox[t]{2mm}{\multirow{4}{*}{\rotatebox[origin=c]{90}{\textbf{Gamma}}}}
      & 0.0 & $ \textbf{83.82}\pm0.3 $           & $ 83.57\pm0.4 $ & $ \textbf{84.10}\pm0.3 $ \\
      & 1.0 & $ 83.71\pm0.4 $           & $ \textbf{83.61}\pm0.6 $ & $ 83.99\pm0.5 $ \\
      & 2.0 & $ 83.38\pm0.3 $           & $ 83.38\pm0.6 $ & $ 83.72\pm0.4 $ \\
      & 5.0 & $ 82.65\pm0.3 $           & $ 82.28\pm0.6 $ & $ 82.68\pm0.5 $ \\
         \toprule 
         \multicolumn{5}{c}{\textbf{HARD Test subset results}}\\
         \toprule
 \parbox[t]{2mm}{\multirow{4}{*}{\rotatebox[origin=c]{90}{\textbf{Gamma}}}}
      & 0.0 & $ \textbf{76.08}\pm0.4 $           & $ 75.63\pm0.7 $ & $ \textbf{76.78}\pm0.4 $ \\
      & 1.0 & $ 75.80\pm0.7 $           & $ \textbf{75.83}\pm0.8 $ & $ 76.69\pm0.9 $ \\
      & 2.0 & $ 75.32\pm0.4 $           & $ 75.37\pm0.9 $ & $ 76.15\pm0.8 $ \\
      & 5.0 & $ 74.35\pm0.5 $           & $ 74.06\pm0.7 $ & $ 74.74\pm0.6 $ \\
         \toprule 
         \multicolumn{5}{c}{\textbf{HANS}}\\
         \toprule
 \parbox[t]{2mm}{\multirow{4}{*}{\rotatebox[origin=c]{90}{\textbf{Gamma}}}}
      & 0.0 & $ 52.20\pm1.0 $           & $ 60.76\pm6.1 $ & $ \textbf{94.23}\pm4.0 $ \\
      & 1.0 & $ 53.61\pm2.0 $           & $ 60.71\pm5.2 $ & $ 88.21\pm3.6 $ \\
      & 2.0 & $ 54.91\pm3.1 $           & $ 57.18\pm3.2 $ & $ 82.69\pm4.3 $ \\
      & 5.0 & $ \textbf{59.54}\pm3.2 $           & $ \textbf{61.46}\pm3.5 $ & $ 67.56\pm4.7 $ \\ \bottomrule
\end{tabular}%
}
\end{minipage}
\end{table}

\section{Results and Discussion}

\subsection{Impact of Focal Loss}
To investigate the effect of focal loss, we ran each set of experiments with 
$6$ different values of the parameter $\gamma \in \{0, 0.5, 1, 2, 5, 10\}$, while keeping all other hyperparameters intact. As mentioned before, $\gamma = 0$ is equivalent to cross-entropy and serves as a baseline. The results summarized in Table \ref{tab:exp1} show that focal loss with $\gamma = 5$ achieves the best performance on the \ac{HANS} test set. However, the accuracy on the datasets' test sets and their hard subsets is generally lowered by increasing $\gamma$.

To further examine the cause of such a behaviour, we plot the distribution of ground truth probabilities of the BERT model on the \ac{MNLI} test set, as shown in Figure \ref{fig:plot_prediction_distribution} (other more detailed figures for this analysis can be found in Appendix \ref{sec:appendix2}). The plot demonstrates the tendency of a focal loss model to produce less certain predictions, a property inherent to its design. Looking back at Figure \ref{fig:focal_loss_gamma}, it is clear that focal loss with higher gamma values rewards even wrongly classified samples, given that the probability of the ground truth label is high enough. In other words, the model will be satisfied with increasing the correct class probability, potentially without producing higher accuracy. Our experiment shows that this phenomenon negatively impacts in-distribution performance and does not improve the classification of harder samples. However, such a design has benefited the models' ability to generalize on out-of-distribution samples, i.e. the \ac{HANS} dataset, meaning that it can also produce less heuristic-prone models and acts as regularization.

The relatively low standard deviations of the models' accuracy on the test sets (as shown in Table \ref{tab:exp1}) suggest that the observed behaviour, its positive and negative aspects, are reliable for the chosen models and datasets.

\begin{figure}[t]
    \centering
    \includegraphics[width=\columnwidth]{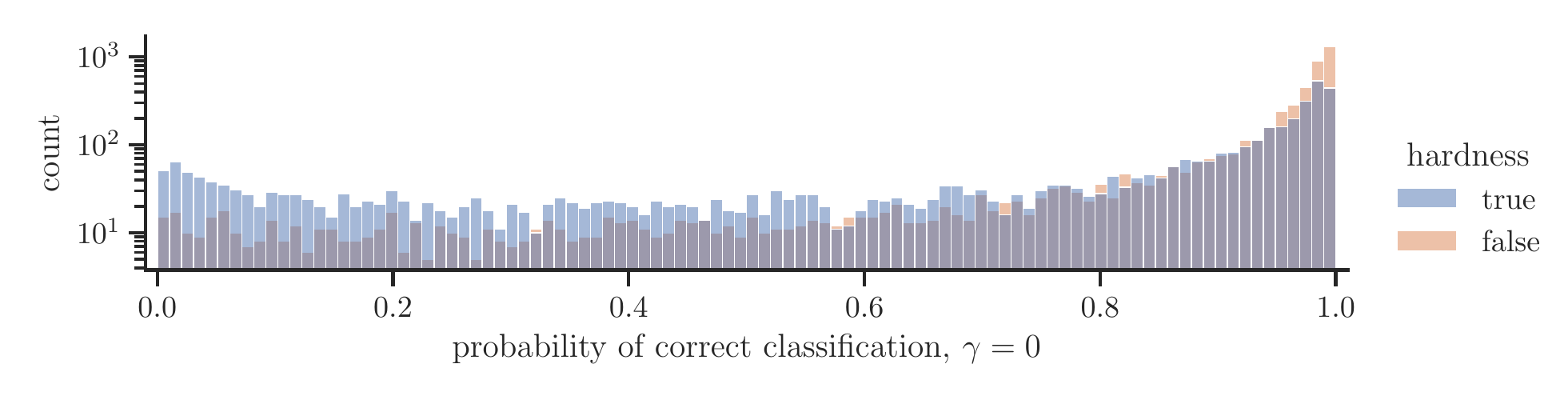}
    \includegraphics[width=\columnwidth]{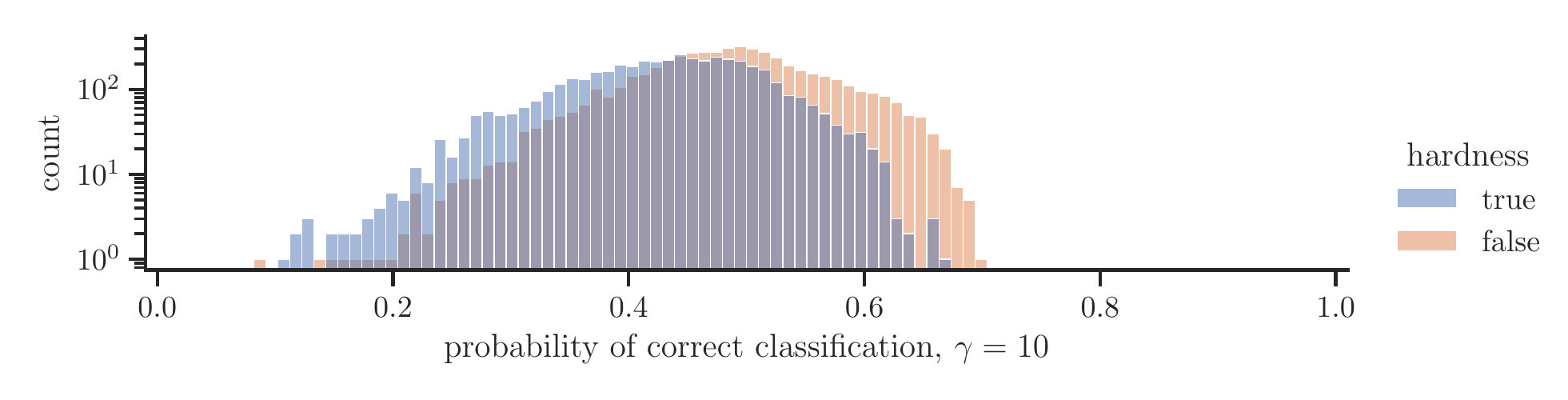}
    \caption{The distribution of ground truth class probabilities of the \ac{MNLI} test set produced by the BERT model for $\gamma = 0$ (cross-entropy) and $\gamma = 10$. Hardness of samples is defined following \cite{gururangan2018annotation}.
    There exists an anticipated trend of hard examples having a lower probability of correct classification.
    The tendency of the probabilities to converge close to $0.5$ when increasing gamma is a direct consequence of the focal loss's design, and can be a positive effect for avoiding learning heuristics by focusing on harder examples, but also removes certainty from correctly classified data.}
    \label{fig:plot_prediction_distribution}
\end{figure}

\subsection{Impact of Focal Loss when Adding HANS Samples to Training}
For the secondary experiment, we observe the influence of focal loss on the BERT model trained on the \ac{MNLI} dataset with added HANS training samples. From Table \ref{tab:exp2}, we can see that with $100$ samples, focal loss still improves, or at least does not reduce performance on the \ac{HANS} test set. However, adding $1000$ samples clearly shows the inefficacy of focal loss when there is a lesser distribution shift between the training set and the test set. This, along with the first experiment, indicates that focal loss on its own is not helpful with learning harder examples in \ac{NLI} tasks, and can likely be useful regularization only for out-of-distribution data. However, it should be noted that the hard classification from \cite{gururangan2018annotation} is not without fault, as it also includes many easily-classified samples. This is visible from the prediction distributions in Figure \ref{fig:plot_prediction_distribution}, as well as the loss distributions in Appendix \ref{sec:appendix2}.

\begin{figure}[t]
    \centering
    \includegraphics[width=\columnwidth]{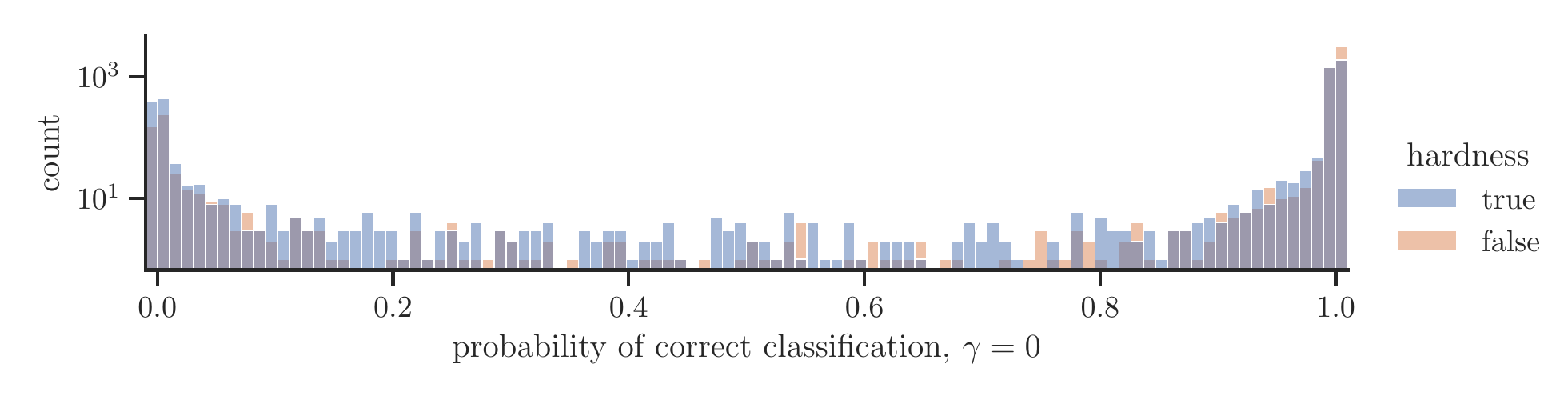}
    \includegraphics[width=\columnwidth]{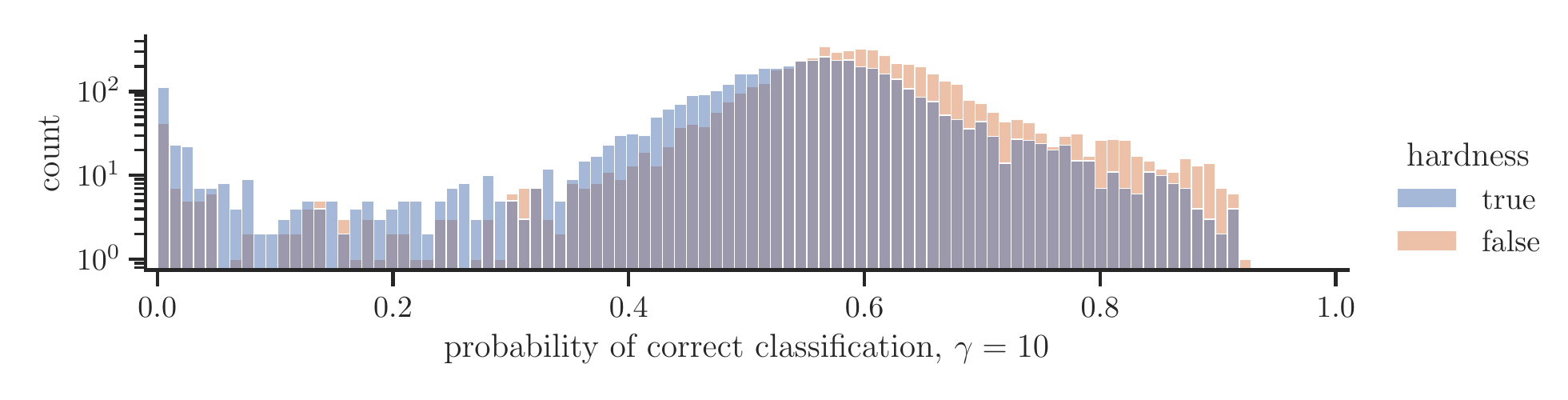}
    \caption{The distribution of ground truth class probabilities of the \ac{MNLI} test set produced by the BERT model for $\gamma = 0$ (cross-entropy) and $\gamma = 10$ when not using early stopping. Unlike the distributions in Figure \ref{fig:plot_prediction_distribution}, here we can see a comparatively more even distribution, with more certain predictions resulting in more correct classifications of hard and previously misclassified samples. }
    \label{fig:plot_prediction_distribution_no_es}
\end{figure}

\subsection{Difficulty of Control: Validation}
\label{difficulty-of-control}
It is worth mentioning that our experiments observe the $\gamma$ parameter separately from other hyperparameters. However, in a real setting, $\gamma$ should be validated. This is likely to prove problematic, as the positive impact of focal loss seems to be mainly visible on out-of-distribution data.

The lack of relevant validation data was especially visible in our early stopping procedure, as we have noticed that the test sets showed better performance in later epochs, which we could not justify using with the available validation data. For example, without early stopping, the BERT model trained on \ac{MNLI} gave an accuracy of $64.76 \pm 0.96$ with cross-entropy and $65.22 \pm 0.81$ with focal loss (for $\gamma=0.5$) on the \ac{HANS} test set, and also showed less deterioration and even some improvement in performance of hard examples from the \ac{MNLI} test set. The problem of validation and hyperparameter tuning has been observed previously \cite{clark2019don} and in this case prevented us from producing results comparable with existing literature \cite{clark2019don,bias-mitigation-2020,utama2020debias,utama2020towards}. We also investigated this difference in performance by plotting the probability distributions as previously in Figure \ref{fig:plot_prediction_distribution}, but without early stopping. This is shown in Figure \ref{fig:plot_prediction_distribution_no_es}. Without early stopping, the probability distribution looks significantly better in so that it produces more certain predictions and leans more heavily to the right (resulting in higher accuracy). This likely means that early stopping underfit the models and prevented them from leveraging the benefits of using focal loss. More broadly, it shows that focal loss on its own is unlikely to work well without a relevant validation set, a possible reason why it performed so well in object detection \cite{focal-loss-in-cv-2017}, and not in our experiments. In line with our second experiment, however, it might also be the case that adding any relevant data to the training set is more beneficial than using it for validating focal loss.

We conclude that focal loss on its own is not good for fighting shallow heuristics and difficult imbalances more generally. Its shortcomings outweigh its positive sides, and other approaches such as debiased focal loss \cite{bias-mitigation-2020} or ensemble learning methods \cite{utama2020towards,sanh2021learning,clark2020learning}, which have shown more promising results (a comparison table can be found in Appendix \ref{sec:appendix2}), should be preferred for \ac{NLI} tasks. Those methods go a step further than just modulating cross-entropy and can effectively discover biases in the dataset, with or without \textit{a priori} knowledge. They improve in-distribution and out-of-distribution performance by controlling the learning process using estimated notions of bias, something that focal loss fails to do on its own.

\section{Conclusion}

In this paper, we investigate the use of focal loss as a means of preventing networks from learning heuristics for the task of \ac{NLI}. We train BERT and InferSent models on the \ac{MNLI} and \ac{SNLI} datasets, and evaluate their generalization performance on the adversary \ac{HANS} dataset. We find that focal loss did help in obtaining better generalization performance with out-of-distribution data (i.e., the \ac{HANS} dataset), but failed in achieving higher accuracy on hard samples. Investigating the ground truth probability distribution of test set samples, we found that focal loss produced models with less certain predictions without managing to learn and improve accuracy by focusing on difficult samples from the training set. Our experiments show that part of the problem is the difficulty of obtaining a relevant validation set, likely causing early stopping to produce underfit models that fail to leverage the benefits of focal loss. Moreover, when a larger number of \ac{HANS} training samples were added to the training set, focal loss had no positive impact on any of the test sets.

An important caveat of our work is that we do not perform hyperparameter search across other model parameters, aside $\gamma$, due to its computational cost, which potentially weakens our claims. However, we believe that the fact that similar trends were observed across repeated experiments on two different models and two different datasets, along with our analysis, is sufficient to support them never the less.

An interesting extension of our work would be to use more expressive models that have been proven to possess better sample efficiency \cite{openAI_scaling_laws} and accuracy on \ac{NLI} tasks \cite{t5}. It is possible that such models could profit more from focal loss and learn harder examples instead of shifting the probabilities to higher values to accommodate the loss function. Towards a similar end, it would also be interesting to investigate the use of different validation sets for early stopping, e.g. using only hard examples for validation. This could prevent models from stopping learning before reaching better performance on harder examples, which is something that is likely to have happened in our experiments. Furthermore, investigating how the HANS, \ac{MNLI}, \ac{SNLI}, and hard subset distributions differ would potentially allow for a better understanding of the problem.

\section{Acknowledgements}
The authors would like to thank Prof. Martin Jaggi for his continued advice and support.
The authors would also like to thank Rabeeh Karimi Mahabadi for her helpful insights.

\clearpage
\bibliography{literature}
\newpage
\clearpage
\appendix

\section{Model Hyperparameters}
\label{sec:appendix1}

\subsection{BERT}
We take a pretrained \texttt{bert-base-uncased} model from HuggingFace with the default configuration (the commit available \href{https://huggingface.co/bert-base-uncased/tree/345fd30026bc3003828be943882dda32ab48b908}{here}) and add a sequence classification head on top (\textit{i.e.}, a linear layer on top of the pooled output). The hyperparameters used for finetuning the model are taken from \cite{mccoy2020berts}, with the distinction of using 10 epochs instead of 3, and performing early stopping on the validation loss:
\begin{itemize}
    \item linear learning rate warmup schedule for 10\% of the optimization steps, followed by a linear decay to $0$
    \item AdamW \cite{adamw} optimizer with $\epsilon=10^{-6}$, $\beta_1=0.9$, and $\beta_2=0.999$
    \item learning rate of $2\times10^{-5}$
    \item weight decay of $0.01$
    \item batch size of 32
    \item gradient clipping by norm, of value $1.0$
    \item 16-bit mixed precision training \cite{mixed}
    \item 10 training epochs
\end{itemize}

\subsection{InferSent}
The hyperparameters of InferSent are taken from \cite{bias-mitigation-2020}. The bidirectional long short-term memory (BiLSTM) encoder has 512 neurons, as does the adjacent nonlinear classifier. For nonlinearity, the $tanh$ function is used. The input words are encoded using 300-dimensional GloVe embeddings. The training is performed for 20 epochs using stochastic gradient descent (SGD), with a starting learning rate of $0.1$. Gradient clipping of norm $5$ is performed. The learning rate has a shrink factor of $5$ that is applied after each epoch. Early stopping is performed based on the validation loss and the parameters with the lowest validation loss are retained for evaluation.

\section{Extra Tables and Figures}
\label{sec:appendix2}

Tables \ref{tab:exp1hans} and \ref{tab:exp2hans} show HANS performance per heuristic type and label. Table \ref{tab:literature-comparison} shows a comparison of our results with the results from literature with the caveat of performing the tuning of $\gamma$ on the test set. As discussed in \ref{difficulty-of-control}, we have no access to an out-of-distribution representative validation set that we could use for hyperparameter tuning. Thus, we follow prior work \cite{clark2019don,bias-mitigation-2020,grand-belinkov-2019-adversarial} and perform model selection on HANS.

\begin{table}
    \centering 
     \caption{
        Comparison of our focal loss results with literature on MNLI matched validation set (used as a test set) and HANS. We report our Cross-Entropy (CE) and best Focal Loss (FL) results with \textbf{no early stopping} performed.
        $\bm{\Delta}$ are the differences with respect to the CE HANS accuracy. Debiased models that were trained with prior knowledge about the heuristics are indicated as \texttt{known-bias}, or as \texttt{self-debias} if they use no prior knowledge about the heuristics biases inherent in the dataset. 
        \ding{163}: results from~\cite{clark2019don}.
        \ding{171}: results from~\cite{bias-mitigation-2020}.
        \ding{117}: results from~\cite{utama2020debias}.
        $\spadesuit$: results from~\cite{utama2020towards}.
        \ding{68}: perform hyper-parameter tuning on the test set.
    }
    \resizebox{0.7\textwidth}{!}{
    \begin{tabular}{llllllllllllll}
        \toprule 
        \textbf{Loss}    &  \textbf{MNLI} & \textbf{HANS}   & \bm{$\Delta$} \\ 
        \toprule 
        
        CE   &   $84.14\pm0.3$  &   $64.76\pm1.0$   &   \\
        
        FL\textsubscript{\textbf{ self-debias}}~\ding{68} & $84.10\pm0.2$ & $65.22\pm0.8$ & $+0.5$ \\ 
        
        \midrule
        
        Learned-Mixin+H~\ding{163}~\ding{68}\textsubscript{ known-bias} & $83.97$ & $66.15$ & $+1.4$ \\
        
        Reweighting~\ding{163}\textsubscript{ known-bias}    & $83.54$ & $69.19$ & $+4.4$\\  
        
        \midrule
        
        PoE~\ding{110}\textsubscript{ known-bias} & $82.9$ &  $67.9$ & $+6.4$\\

        \midrule 
        
        PoE~\ding{171}\textsubscript{ known-bias}               & $84.19$  & $66.31\pm0.6$  & $+1.6$\\ 
        
        DFL~\ding{171}\textsubscript{ known-bias}  &  $83.95$ & $69.26\pm0.2$ & $+4.5$\\
        
        DFL~\ding{171}~\ding{68}\textsubscript{ known-bias}  &  $82.76$ & $\bm{71.95}\pm1.4$ & $\bm{+7.2}$  \\
        
        \midrule
        
        Conf-reg~\ding{117}\textsubscript{ known-bias} & $84.5$ & $69.1$ & $+4.3$ \\
        
        Conf-reg $\spadesuit$\textsubscript{ \textbf{self-debias}} & $83.9$ & $67.7$ & $+2.9$\\
        
        Reweighting $\spadesuit$\textsubscript{ \textbf{self-debias}} & $82.3$ &  $69.7$ & $+4.9$ \\
        
        \bottomrule 
    \end{tabular}}
    \label{tab:literature-comparison} 
\end{table}
\begin{table}
\centering
\caption{HANS per-heuristic acccuracies of the BERT and InferSent models trained on the \ac{MNLI} and \ac{SNLI} datasets. For \ac{MNLI}, we have used the mismatched validation subset for early stopping. We report the average accuracy across 5 runs alongside the corresponding standard deviation.}
\label{tab:exp1hans}
\resizebox{\columnwidth}{!}{%
\begin{tabular}{@{}cccccccccc@{}}
\toprule 
&& \multicolumn{4}{c}{\textbf{MNLI}} & \multicolumn{4}{c}{\textbf{SNLI}} \\
\cmidrule(r){3-6} \cmidrule(l){7-10} 
&& \multicolumn{2}{c}{\textbf{BERT}} & \multicolumn{2}{c}{\textbf{InferSent}} & \multicolumn{2}{c}{\textbf{BERT}} & \multicolumn{2}{c}{\textbf{InferSent}} \\
&& Entailed & Non-entailed & Entailed & Non-entailed & Entailed & Non-entailed & Entailed & Non-entailed \\
\toprule 
\multicolumn{10}{c}{\textbf{Lexical Overlap}}\\
\toprule
\parbox[t]{2mm}{\multirow{6}{*}{\rotatebox[origin=c]{90}{\textbf{Gamma}}}}
      & 0.0  & $ \textbf{99.06}\pm0.4 $  & $ 7.21\pm3.4 $   & $ \textbf{98.82}\pm0.7 $ & $ 1.23\pm0.6 $  & $ 98.74\pm0.8 $  & $ 44.81\pm10.4 $ & $ \textbf{99.03}\pm0.6 $  & $ 2.45\pm1.1 $   \\
      & 0.5  & $ 98.58\pm1.2 $  & $ 10.33\pm5.3 $  & $ 98.07\pm1.5 $ & $ 2.00\pm1.5 $  & $ \textbf{99.19}\pm0.4 $  & $ 40.27\pm8.6 $  & $ 97.34\pm1.5 $  & $ 5.72\pm3.4 $   \\
      & 1.0  & $ 98.20\pm2.0 $  & $ 11.44\pm7.3 $  & $ 97.88\pm0.8 $ & $ 2.58\pm1.0 $  & $ 99.02\pm0.4 $  & $ 39.83\pm15.2 $ & $ 95.64\pm1.4 $  & $ 7.01\pm1.6 $   \\
      & 2.0  & $ 97.62\pm1.2 $  & $ 15.78\pm8.7 $  & $ 96.52\pm2.3 $ & $ 5.19\pm3.4 $  & $ 98.16\pm1.1 $  & $ 53.35\pm8.0 $  & $ 92.31\pm4.2 $  & $ 11.14\pm4.9 $  \\
      & 5.0  & $ 90.90\pm4.8 $  & $ 23.89\pm6.4 $  & $ 77.28\pm5.1 $ & $ 32.24\pm7.2 $ & $ 89.60\pm8.0 $  & $ 70.79\pm10.9 $ & $ 54.05\pm16.3 $ & $ 48.48\pm16.8 $ \\
      & 10.0 & $ 15.21\pm19.9 $ & $ \textbf{90.50}\pm14.1 $ & $ 0.08\pm0.1 $  & $ \textbf{99.80}\pm0.3 $ & $ 15.95\pm34.3 $ & $ \textbf{97.74}\pm4.7 $  & $ 1.17\pm0.5 $   & $ \textbf{98.61}\pm0.4 $  \\
\toprule 
\multicolumn{10}{c}{\textbf{Subsequence}}\\
\toprule
\parbox[t]{2mm}{\multirow{6}{*}{\rotatebox[origin=c]{90}{\textbf{Gamma}}}}
      & 0.0  & $ 99.99\pm0.0 $  & $ 0.84\pm0.4 $   & $ \textbf{99.39}\pm0.2 $ & $ 2.00\pm1.2 $  & $ \textbf{99.96}\pm0.1 $  & $ 3.52\pm3.5 $   & $ \textbf{99.06}\pm0.5 $  & $ 1.12\pm0.8 $   \\
      & 0.5  & $ \textbf{100.00}\pm0.0 $ & $ 1.51\pm1.1 $   & $ 98.77\pm1.3 $ & $ 3.64\pm2.6 $  & $ 99.83\pm0.4 $  & $ 2.81\pm2.4 $   & $ 98.70\pm0.4 $  & $ 2.31\pm1.4 $   \\
      & 1.0  & $ 99.97\pm0.0 $  & $ 2.30\pm1.8 $   & $ 98.18\pm1.0 $ & $ 3.80\pm1.5 $  & $ \textbf{99.96}\pm0.1 $  & $ 2.29\pm1.5 $   & $ 98.35\pm0.9 $  & $ 3.95\pm1.2 $   \\
      & 2.0  & $ 99.95\pm0.1 $  & $ 4.84\pm3.4 $   & $ 97.44\pm1.9 $ & $ 5.52\pm3.7 $  & $ 99.86\pm0.2 $  & $ 4.81\pm2.7 $   & $ 96.14\pm2.7 $  & $ 5.94\pm2.7 $   \\
      & 5.0  & $ 95.73\pm5.0 $  & $ 21.30\pm16.2 $ & $ 74.30\pm8.0 $ & $ 30.69\pm4.1 $ & $ 97.08\pm3.6 $ & $ 16.76\pm10.2 $ & $ 77.47\pm10.4 $ & $ 30.45\pm12.8 $ \\
      & 10.0 & $ 20.68\pm22.4 $ & $ \textbf{93.81}\pm10.9 $ & $ 0.75\pm0.9 $  & $ \textbf{98.63}\pm1.4 $ & $ 47.32\pm31.9 $ & $ \textbf{82.86}\pm27.5 $ & $ 6.70\pm2.2 $   & $ \textbf{93.81}\pm2.6 $  \\
\toprule 
\multicolumn{10}{c}{\textbf{Constituent}}\\
\toprule
\parbox[t]{2mm}{\multirow{6}{*}{\rotatebox[origin=c]{90}{\textbf{Gamma}}}}
      & 0.0  & $ \textbf{99.77}\pm0.2 $  & $ 6.36\pm4.3 $   & $ \textbf{97.98}\pm1.4 $ & $ 4.69\pm2.0 $  & $ 99.68\pm0.5 $  & $ 1.78\pm1.3 $   & $ \textbf{99.07}\pm0.5 $  & $ 2.95\pm1.9 $   \\
      & 0.5  & $ 99.76\pm0.2 $  & $ 6.73\pm5.0 $   & $ 96.96\pm2.8 $ & $ 6.22\pm3.8 $  & $ \textbf{99.90}\pm0.1 $  & $ 1.03\pm1.0 $   & $ 97.71\pm0.6 $  & $ 5.09\pm1.9 $   \\
      & 1.0  & $ 99.48\pm0.4 $  & $ 10.26\pm7.2 $  & $ 96.39\pm2.1 $ & $ 7.29\pm2.6 $  & $ 99.65\pm0.4 $  & $ 1.18\pm1.6 $   & $ 97.33\pm0.8 $  & $ 5.83\pm1.3 $   \\
      & 2.0  & $ 98.29\pm1.0 $  & $ 13.00\pm8.9 $  & $ 94.34\pm3.3 $ & $ 9.87\pm4.4 $  & $ 99.81\pm0.4 $  & $ 0.96\pm0.6 $   & $ 94.84\pm2.9 $  & $ 9.12\pm2.8 $   \\
      & 5.0  & $ 89.41\pm8.0 $  & $ 36.02\pm11.6 $ & $ 79.10\pm4.1 $ & $ 25.28\pm4.2 $ & $ 95.98\pm4.6 $  & $ 9.76\pm11.5 $  & $ 71.42\pm12.5 $ & $ 32.97\pm13.1 $ \\
      & 10.0 & $ 23.46\pm24.8 $ & $ \textbf{93.66}\pm8.6 $  & $ 2.30\pm2.0 $  & $ \textbf{97.51}\pm1.7 $ & $ 30.90\pm35.8 $ & $ \textbf{77.48}\pm34.7 $ & $ 8.43\pm2.6 $   & $ \textbf{91.22}\pm2.3 $ \\
\toprule 
\multicolumn{10}{c}{\textbf{Overall Accuracy}}\\
\toprule
 \parbox[t]{2mm}{\multirow{6}{*}{\rotatebox[origin=c]{90}{\textbf{Gamma}}}}
     & 0.0   & \multicolumn{2}{c}{$ 52.20\pm1.0 $} & \multicolumn{2}{c}{$ 50.68\pm0.3 $} & \multicolumn{2}{c}{$ 58.08\pm2.3 $} & \multicolumn{2}{c}{$ 50.61\pm0.4 $} \\
     & 0.5   & \multicolumn{2}{c}{$ 52.82\pm1.3 $} & \multicolumn{2}{c}{$ 50.94\pm0.4 $} & \multicolumn{2}{c}{$ 57.17\pm1.5 $} & \multicolumn{2}{c}{$ 51.14\pm0.6 $} \\
     & 1.0   & \multicolumn{2}{c}{$ 53.61\pm2.0 $} & \multicolumn{2}{c}{$ 51.02\pm0.2 $} & \multicolumn{2}{c}{$ 56.99\pm2.8 $} & \multicolumn{2}{c}{$ 51.35\pm0.3 $} \\
     & 2.0   & \multicolumn{2}{c}{$ 54.91\pm3.1 $} & \multicolumn{2}{c}{$ 51.48\pm0.7 $} & \multicolumn{2}{c}{$ 59.49\pm1.5 $} & \multicolumn{2}{c}{$ 51.58\pm0.3 $} \\
     & 5.0   & \multicolumn{2}{c}{$ \textbf{59.54}\pm3.2 $} & \multicolumn{2}{c}{$ \textbf{53.15}\pm0.7 $} & \multicolumn{2}{c}{$ \textbf{63.33}\pm2.8 $} & \multicolumn{2}{c}{$ \textbf{52.47}\pm0.7 $} \\
     & 10.0  & \multicolumn{2}{c}{$ 56.22\pm5.8 $} & \multicolumn{2}{c}{$ 49.85\pm0.2 $} & \multicolumn{2}{c}{$ 58.71\pm6.0 $} & \multicolumn{2}{c}{$ 49.99\pm0.3 $} \\
\bottomrule
\end{tabular}%
}
\end{table}
\begin{table}
\centering
\caption{HANS per-heuristic accuracies of the BERT model trained on \ac{MNLI} with random \ac{HANS} training samples added to the training test. MNLI mismatched validation set was used for early stopping. We report the average accuracy across 5 runs alongside the corresponding standard deviation.}
\label{tab:exp2hans}

\resizebox{0.8\columnwidth}{!}{%
\begin{tabular}{@{}cccccccc@{}}
\toprule
\multicolumn{8}{c}{\textbf{BERT on MNLI w/ HANS augmentation}} \\  
\toprule
&& \multicolumn{6}{c}{\textbf{\#HANS examples added to train}} \\  
&& \multicolumn{2}{c}{$\textbf{0}$} & \multicolumn{2}{c}{$\textbf{100}$} & \multicolumn{2}{c}{$\textbf{1000}$} \\
&& Entailed & Non-entailed & Entailed & Non-entailed & Entailed & Non-entailed \\
\toprule 
\multicolumn{8}{c}{\textbf{Lexical Overlap}}\\
\toprule
\parbox[t]{2mm}{\multirow{4}{*}{\rotatebox[origin=c]{90}{\textbf{Gamma}}}}
      & 0.0 & $ \textbf{99.06}\pm0.4 $ & $ 7.21\pm3.4 $   & $ 92.30\pm9.1 $  & $ 33.29\pm24.3 $ & $ \textbf{89.75}\pm6.0 $  & $ \textbf{95.55}\pm5.3 $  \\
      & 1.0 & $ 98.20\pm2.0 $ & $ 11.44\pm7.3 $  & $ 94.18\pm4.8 $  & $ 33.64\pm17.1 $ & $ 82.98\pm10.5 $ & $ 94.49\pm5.8 $  \\
      & 2.0 & $ 97.62\pm1.2 $ & $ 15.78\pm8.7 $  & $ \textbf{95.93}\pm2.8 $  & $ 22.12\pm13.8 $ & $ 83.21\pm7.6 $  & $ 86.00\pm14.9 $ \\
      & 5.0 & $ 90.90\pm4.8 $ & $ \textbf{23.89}\pm6.4 $  & $ 82.39\pm15.2 $ & $ \textbf{40.06}\pm16.1 $ & $ 82.66\pm10.7 $ & $ 46.44\pm20.5 $ \\

\toprule 
\multicolumn{8}{c}{\textbf{Subsequence}}\\
\toprule
\parbox[t]{2mm}{\multirow{4}{*}{\rotatebox[origin=c]{90}{\textbf{Gamma}}}}
      & 0.0 & $ \textbf{99.99}\pm0.0 $ & $ 0.84\pm0.4 $ & $ 97.71\pm3.5 $ & $ 16.55\pm18.1 $ & $ \textbf{97.00}\pm2.0 $ & $ \textbf{91.16}\pm10.5 $ \\
      & 1.0 & $ 99.97\pm0.0 $ & $ 2.30\pm1.8 $   & $ 97.81\pm2.3 $  & $ 16.36\pm13.4 $ & $ 93.86\pm3.7 $  & $ 79.60\pm13.9 $ \\
      & 2.0 & $ 99.95\pm0.1 $ & $ 4.84\pm3.4 $   & $ \textbf{99.35}\pm0.5 $  & $ 8.84\pm7.4 $   & $ 94.56\pm4.0 $  & $ 67.01\pm16.0 $ \\
      & 5.0 & $ 95.73\pm5.0 $ & $ \textbf{21.30}\pm16.2 $ & $ 89.00\pm17.9 $ & $ \textbf{29.39}\pm22.6 $ & $ 92.64\pm6.7 $  & $ 44.70\pm20.8 $ \\

\toprule 
\multicolumn{8}{c}{\textbf{Constituent}}\\
\toprule
\parbox[t]{2mm}{\multirow{4}{*}{\rotatebox[origin=c]{90}{\textbf{Gamma}}}}
      & 0.0 & $ \textbf{99.77}\pm0.2 $ & $ 6.36\pm4.3 $   & $ 95.52\pm5.5 $  & $ 29.20\pm14.0 $ & $ \textbf{95.53}\pm1.2 $  & $ \textbf{96.36}\pm4.4 $  \\
      & 1.0 & $ 99.48\pm0.4 $ & $ 10.26\pm7.2 $  & $ 95.30\pm3.1 $  & $ 26.94\pm13.9 $ & $ 91.97\pm2.9 $  & $ 86.34\pm7.1 $  \\
      & 2.0 & $ 98.29\pm1.0 $ & $ 13.00\pm8.9 $  & $ \textbf{98.08}\pm1.5 $  & $ 18.75\pm7.8 $  & $ 87.85\pm7.9 $  & $ 77.53\pm8.9 $  \\
      & 5.0 & $ 89.41\pm8.0 $ & $ \textbf{36.02}\pm11.6 $ & $ 85.35\pm09.97 $ & $ \textbf{42.57}\pm18.32 $ & $ 87.00\pm7.36 $  & $ 51.92\pm11.75 $ \\

\toprule 
\multicolumn{8}{c}{\textbf{Overall Accuracy}}\\
\toprule
 \parbox[t]{2mm}{\multirow{4}{*}{\rotatebox[origin=c]{90}{\textbf{Gamma}}}}
      & 0.0 & \multicolumn{2}{c}{$ 52.20\pm1.0 $}           & \multicolumn{2}{c}{$ 60.76\pm6.1 $} & \multicolumn{2}{c}{$ \textbf{94.23}\pm4.0 $} \\
      & 1.0 & \multicolumn{2}{c}{$ 53.61\pm2.0 $}           & \multicolumn{2}{c}{$ 60.71\pm5.2 $} & \multicolumn{2}{c}{$ 88.21\pm3.6 $} \\
      & 2.0 & \multicolumn{2}{c}{$ 54.91\pm3.1 $}           & \multicolumn{2}{c}{$ 57.18\pm3.2 $} & \multicolumn{2}{c}{$ 82.69\pm4.3 $} \\
      & 5.0 & \multicolumn{2}{c}{$ \textbf{59.54}\pm3.2 $}           & \multicolumn{2}{c}{$ \textbf{61.46}\pm3.5 $} & \multicolumn{2}{c}{$ 67.56\pm4.7 $} \\ \bottomrule

\end{tabular}%
}
\end{table}

\begin{figure}
    \centering
    \includegraphics[width=0.7\columnwidth]{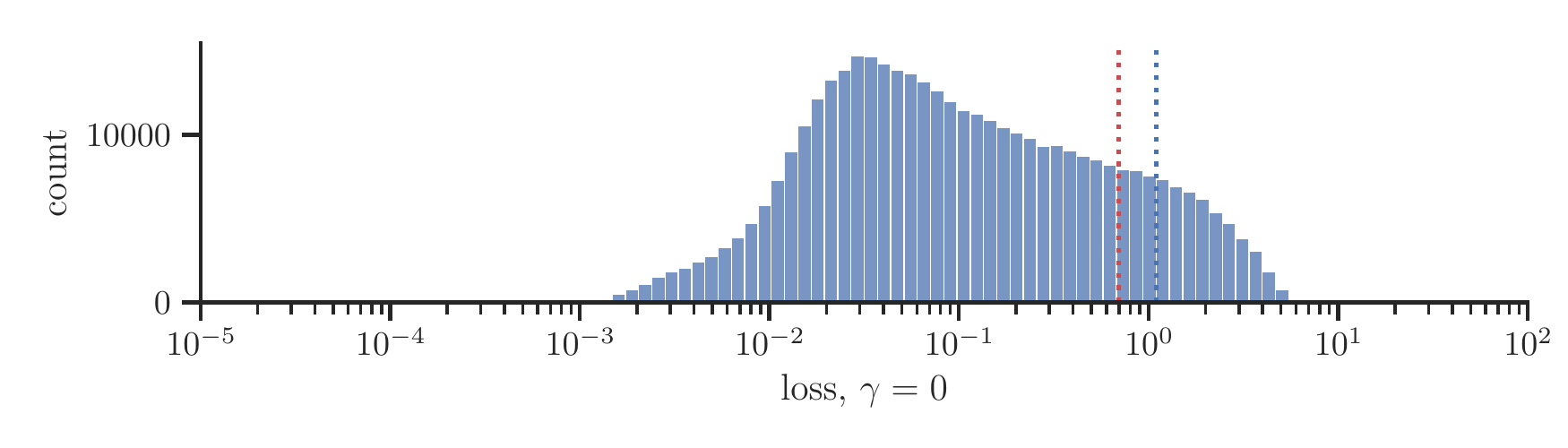}
    \includegraphics[width=0.7\columnwidth]{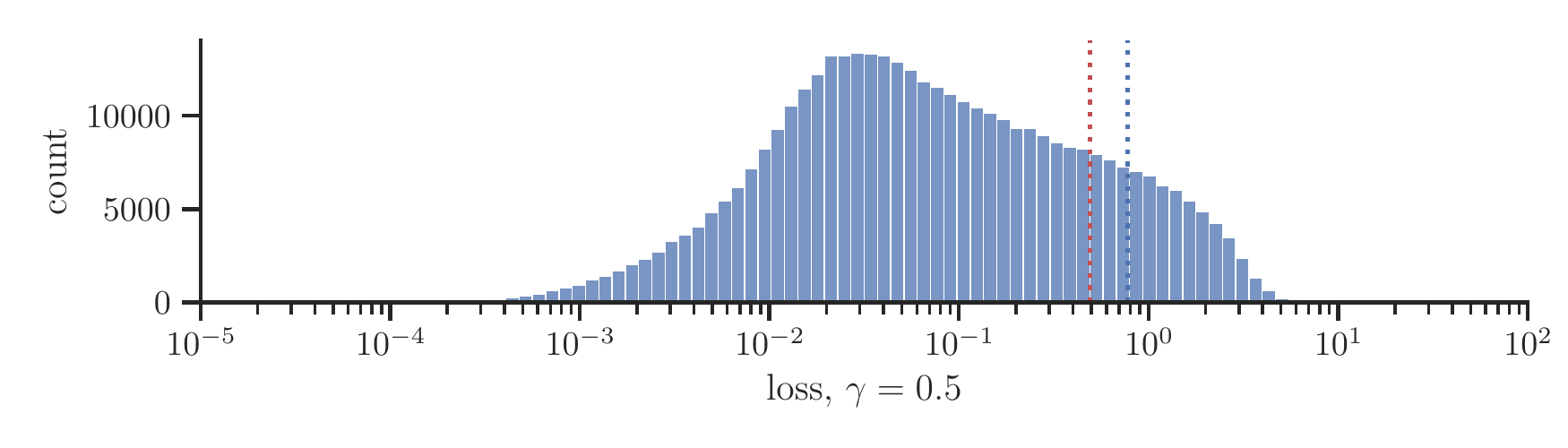}
    \includegraphics[width=0.7\columnwidth]{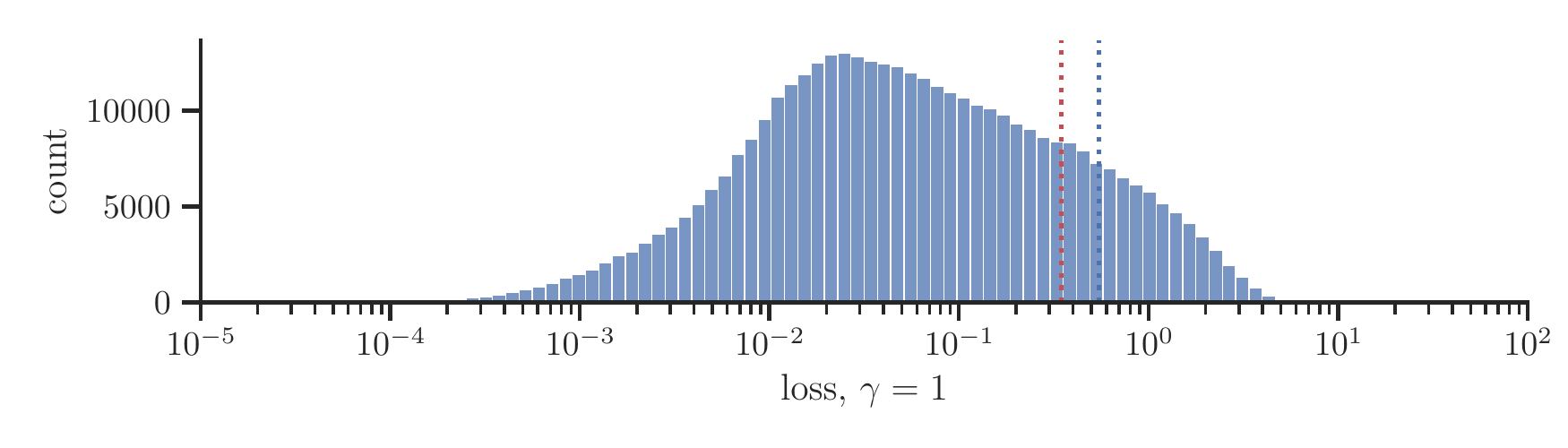}
    \includegraphics[width=0.7\columnwidth]{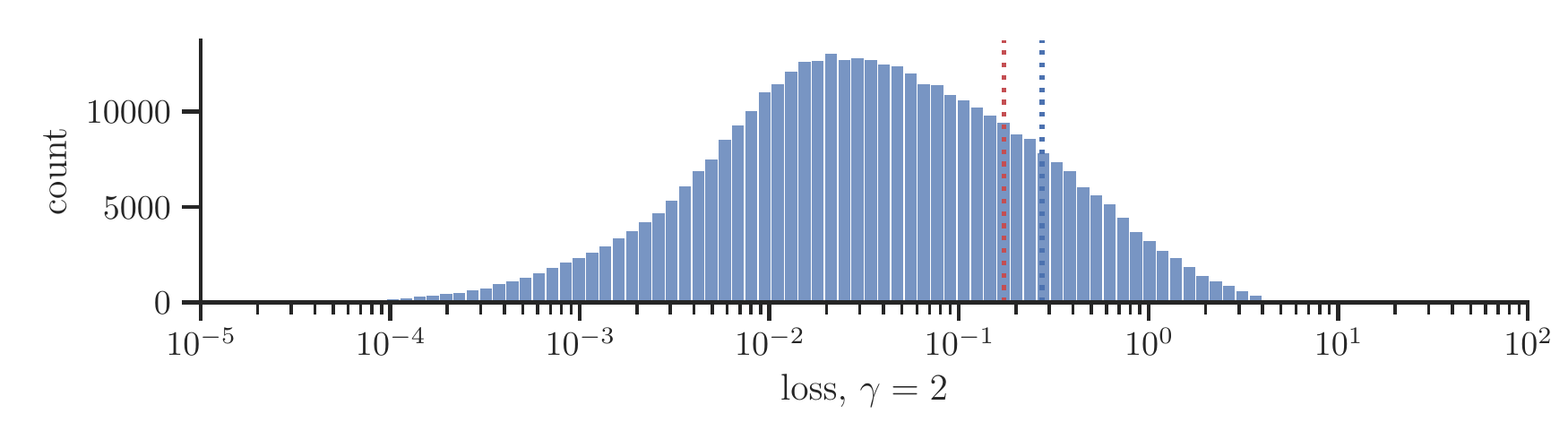}
    \includegraphics[width=0.7\columnwidth]{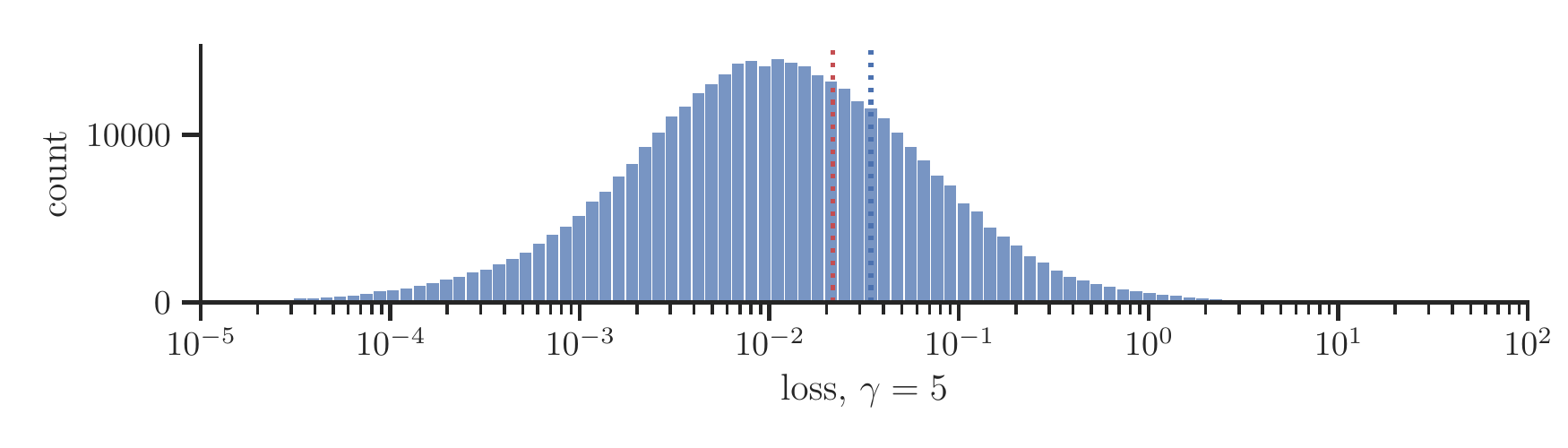}
    \includegraphics[width=0.7\columnwidth]{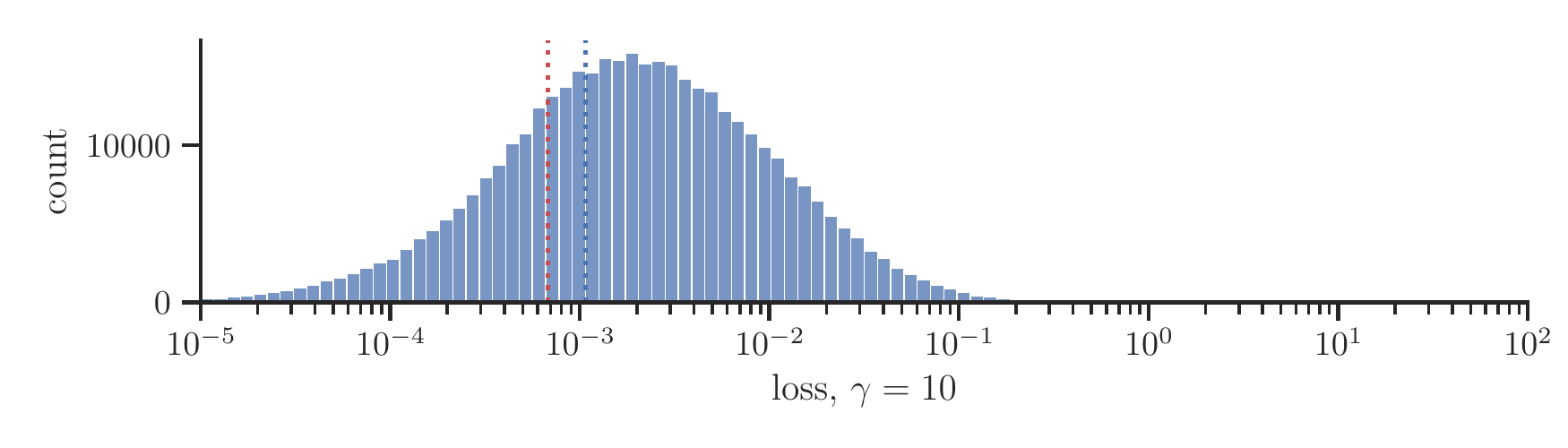}
    \caption{
    The distribution of loss of the \ac{MNLI} train set produced by the BERT model for increasing values of $\gamma$. Hardness of samples is defined following \cite{gururangan2018annotation}. The blue line designates the loss for $0.33$ probability given to the ground truth class, which is the lowest possible probability for correct classification with three-label problems. Similarly, the red line designates the $0.5$ correct class prediction probability, the probability at which correct classification is guaranteed.
    }
    \label{fig:plot-a-mnli-train-v1}
\end{figure}

\begin{figure}
    \centering
    \includegraphics[width=0.75\columnwidth]{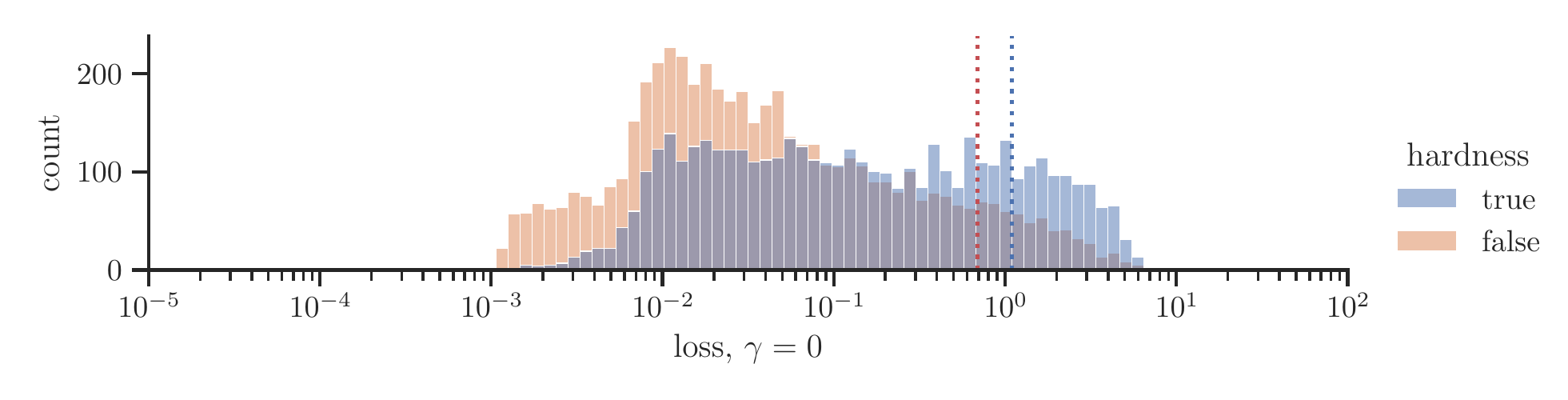}
    \includegraphics[width=0.75\columnwidth]{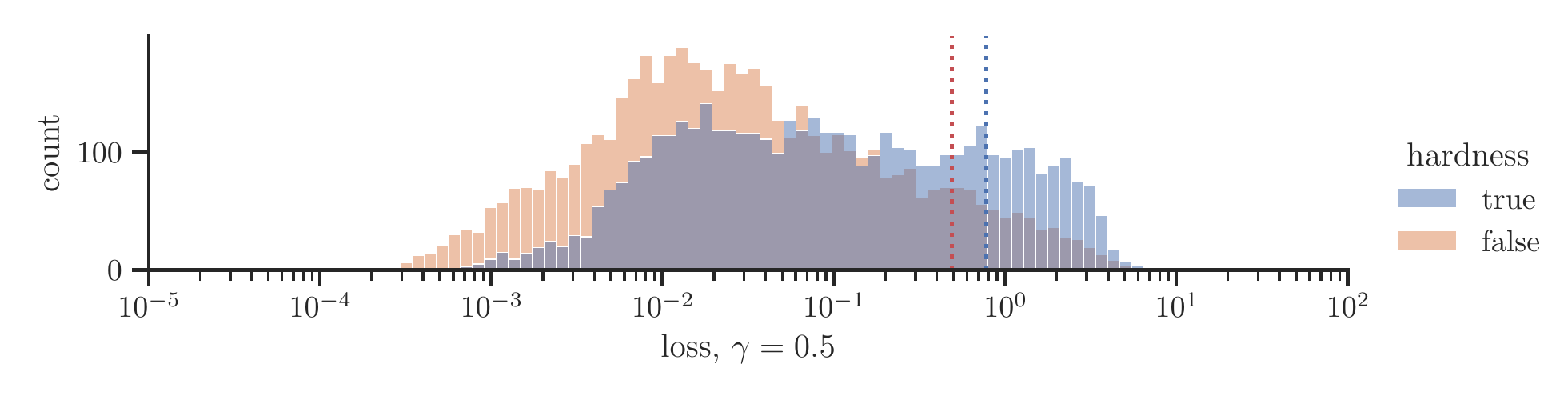}
    \includegraphics[width=0.75\columnwidth]{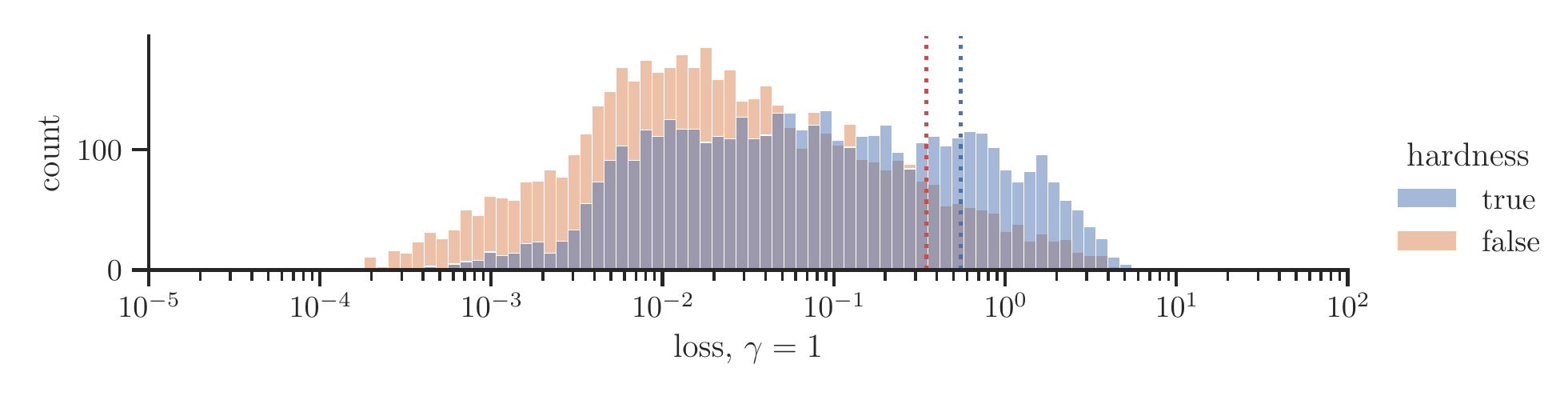}
    \includegraphics[width=0.75\columnwidth]{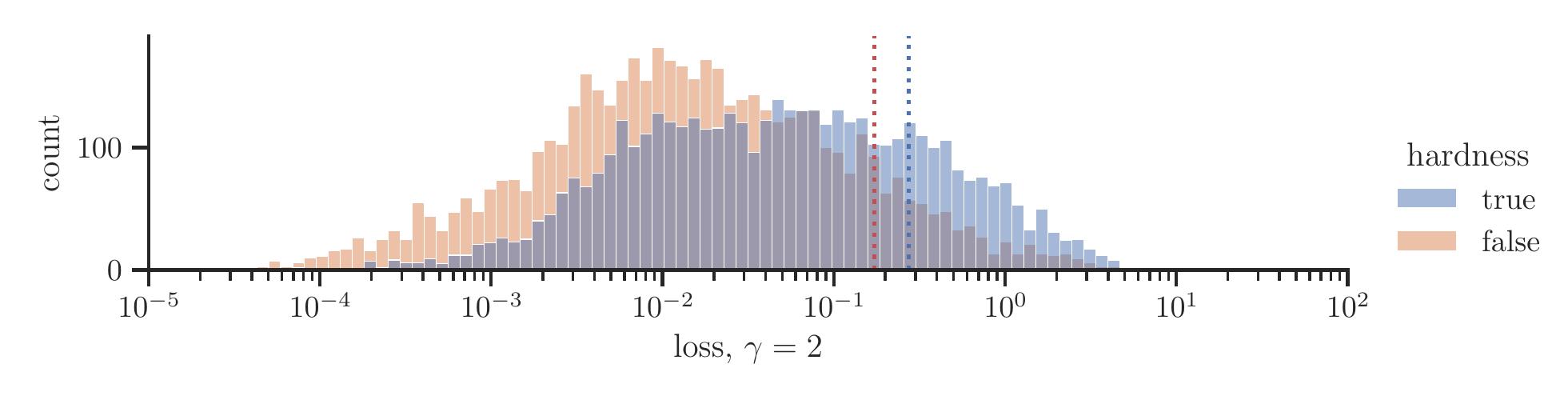}
    \includegraphics[width=0.75\columnwidth]{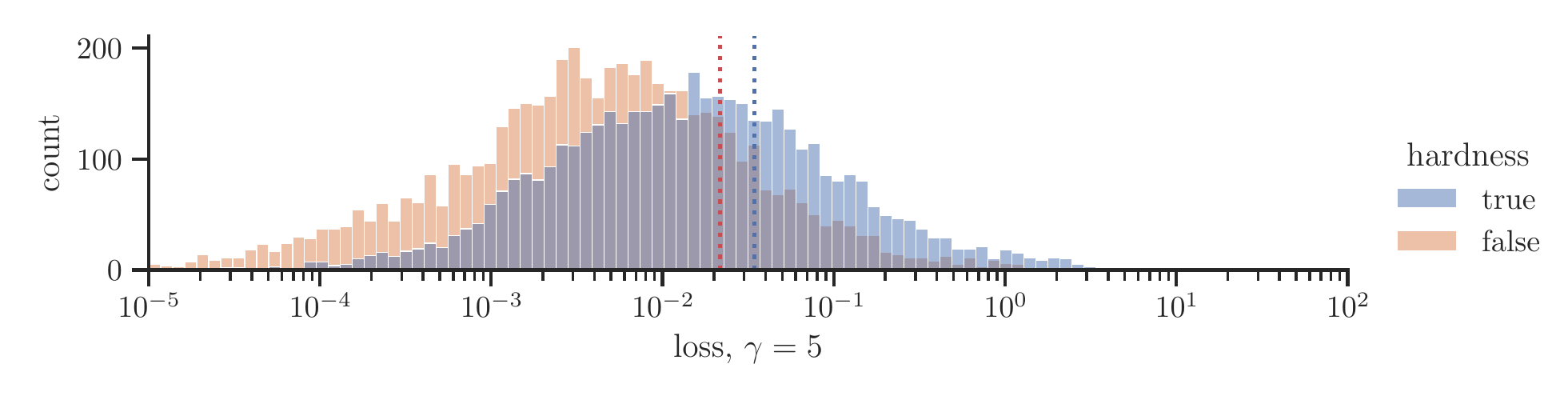}
    \includegraphics[width=0.75\columnwidth]{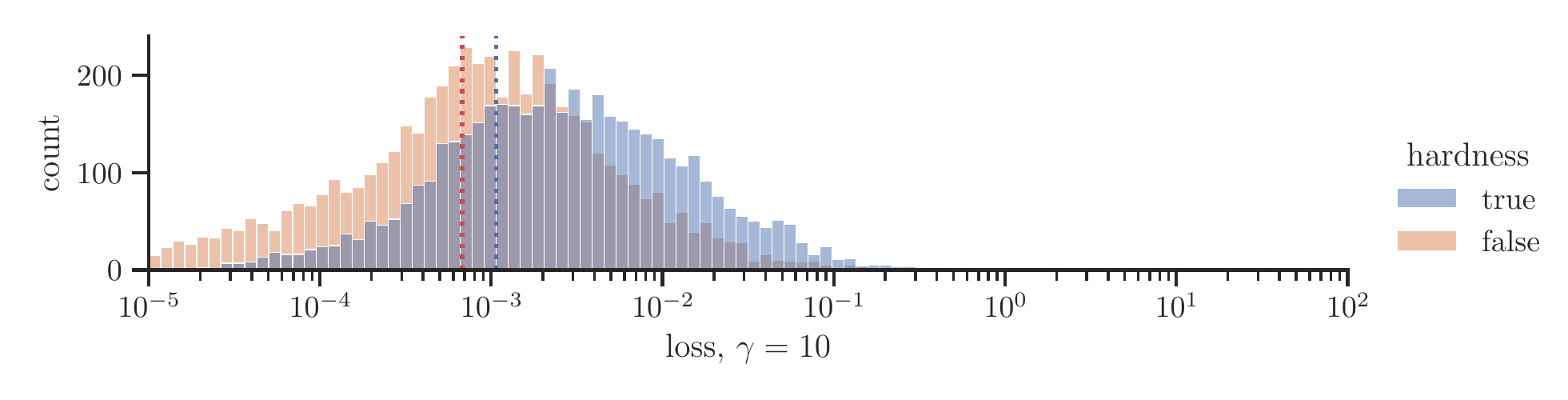}
    \caption{
    The distribution of loss of the \ac{MNLI} test set produced by the BERT model for increasing values of $\gamma$. Hardness of samples is defined following \cite{gururangan2018annotation}. The blue line designates the loss for $0.33$ probability given to the ground truth class, which is the lowest possible probability for correct classification with three-label problems. Similarly, the red line designates the $0.5$ correct class prediction probability, the probability at which correct classification is guaranteed.
    }
    \label{fig:plot-a-mnli-valid-v1}
\end{figure}

\begin{figure}
    \centering
    \includegraphics[width=0.75\columnwidth]{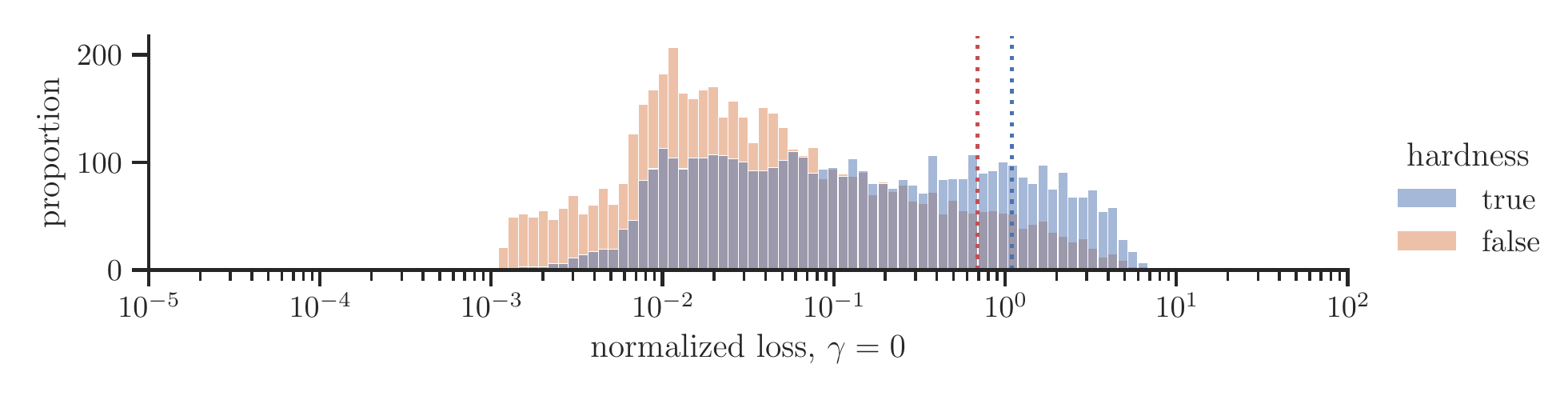}
    \includegraphics[width=0.75\columnwidth]{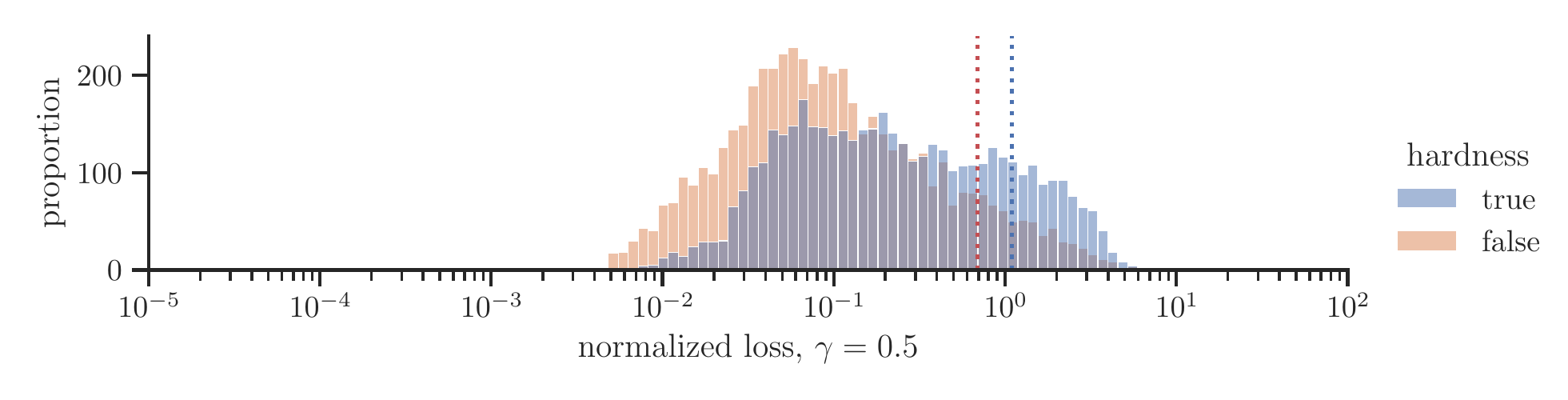}
    \includegraphics[width=0.75\columnwidth]{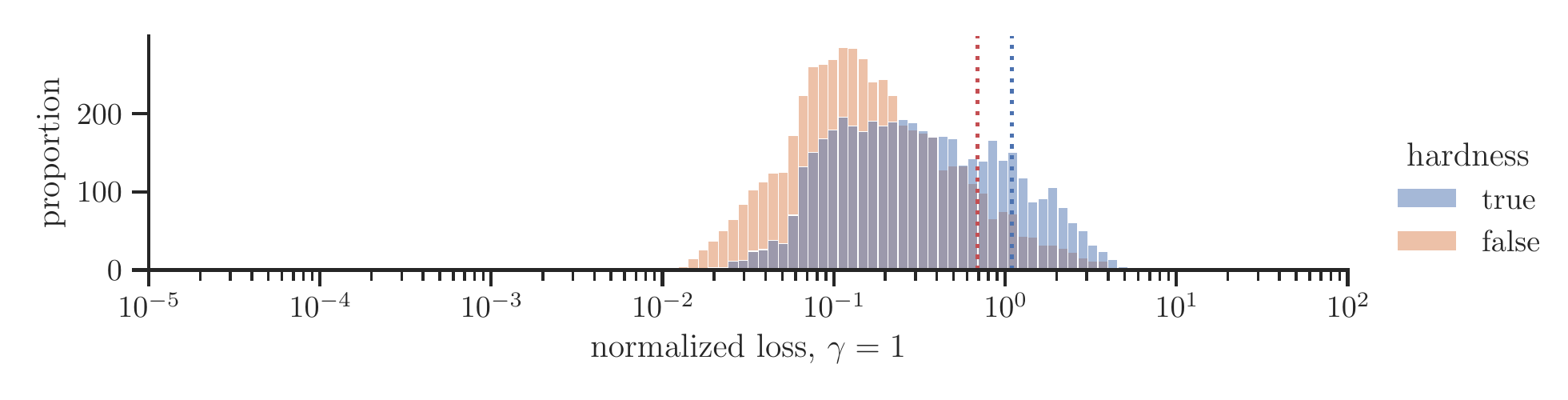}
    \includegraphics[width=0.75\columnwidth]{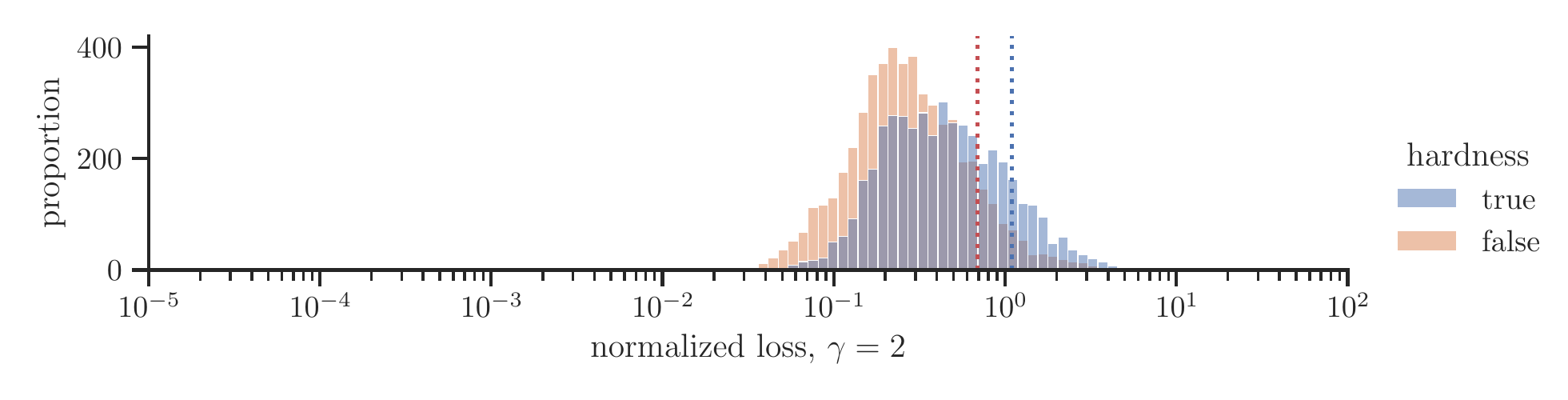}
    \includegraphics[width=0.75\columnwidth]{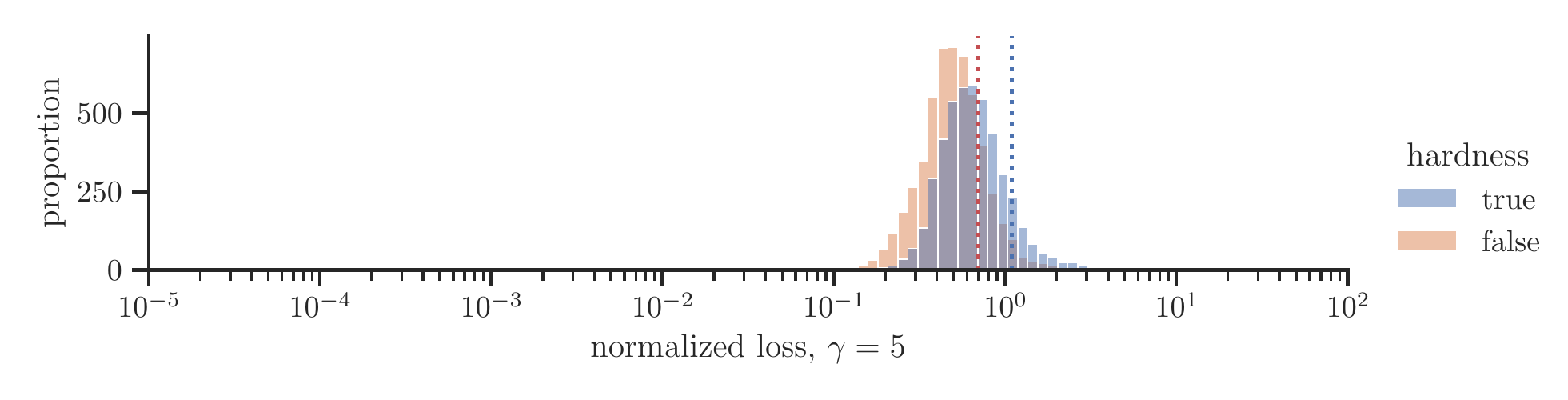}
    \includegraphics[width=0.75\columnwidth]{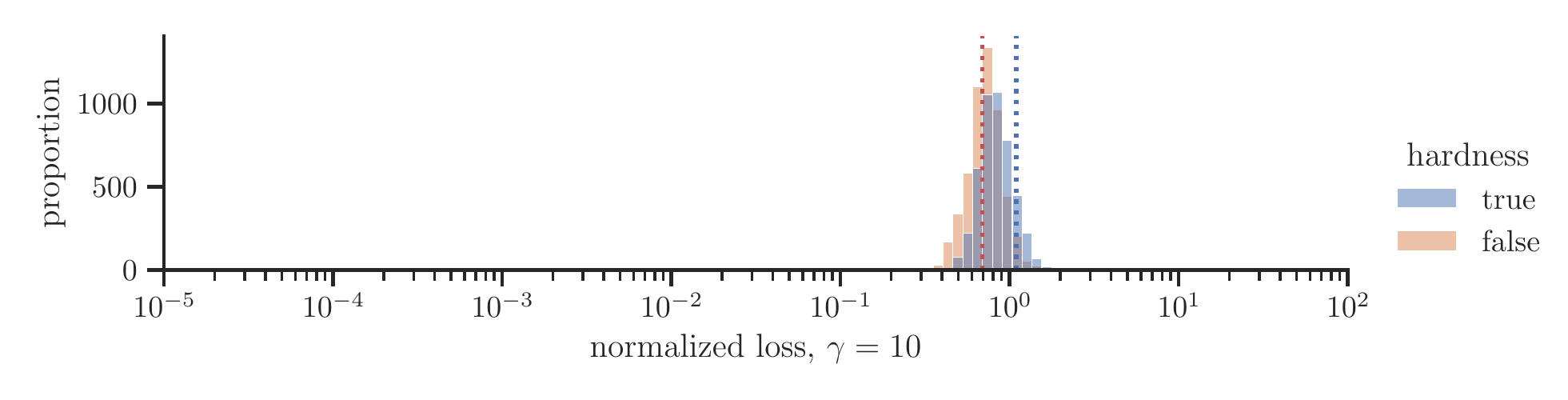}
    \caption{
    The distribution of normalized loss of the \ac{MNLI} test set produced by the BERT model for increasing values of $\gamma$. We normalize the loss by computing the cross-entropy loss based on the outputted class prediction probabilities. Hardness of samples is defined following \cite{gururangan2018annotation}. The blue line designates the loss for $0.33$ probability given to the ground truth class, which is the lowest possible probability for correct classification with three-label problems. Similarly, the red line designates the $0.5$ correct class prediction probability, the probability at which correct classification is guaranteed.
    }
    \label{fig:plot-a-mnli-valid-v2}
\end{figure}

\begin{figure}
    \centering
    \includegraphics[width=0.75\columnwidth]{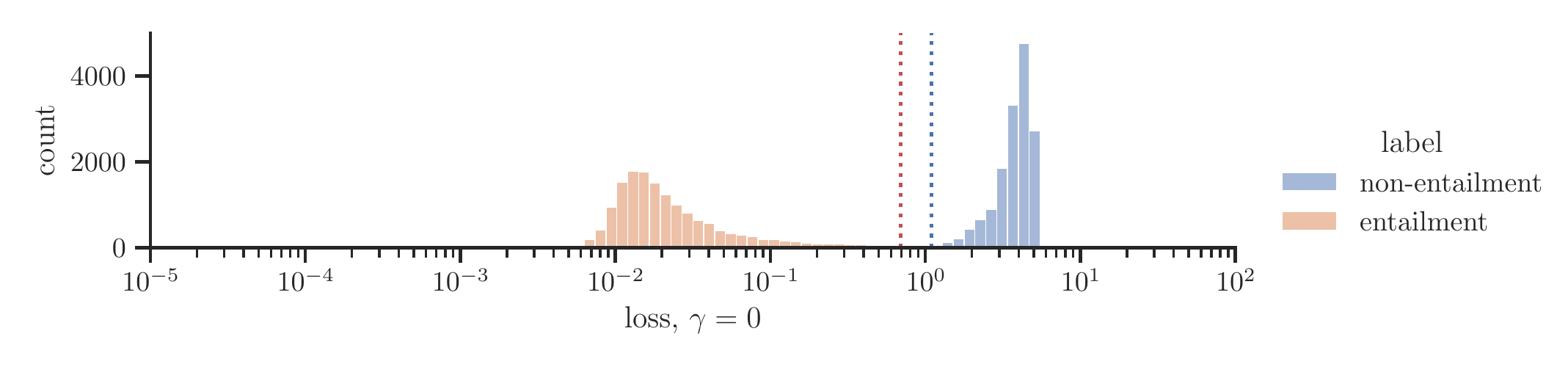}
    \includegraphics[width=0.75\columnwidth]{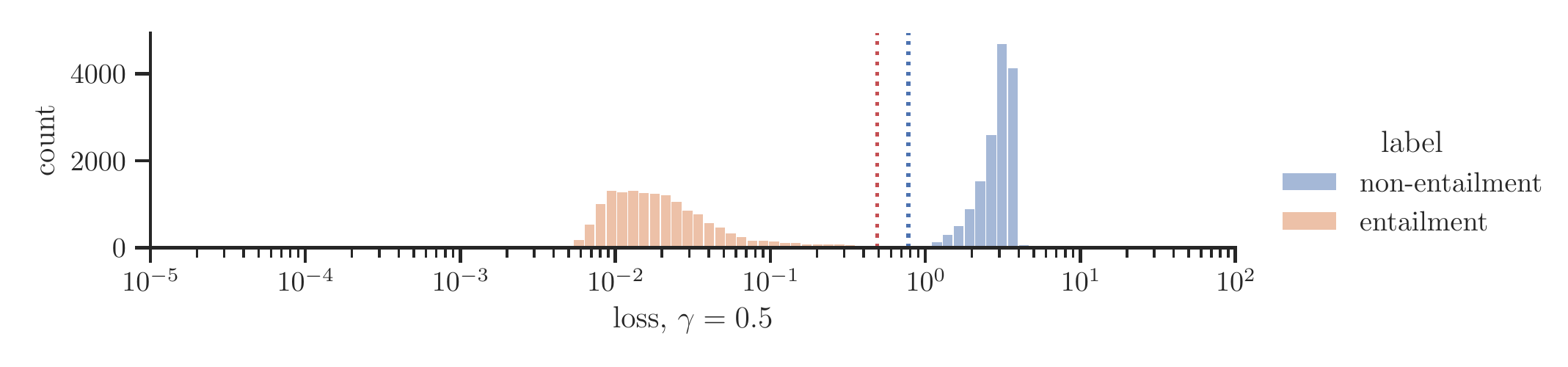}
    \includegraphics[width=0.75\columnwidth]{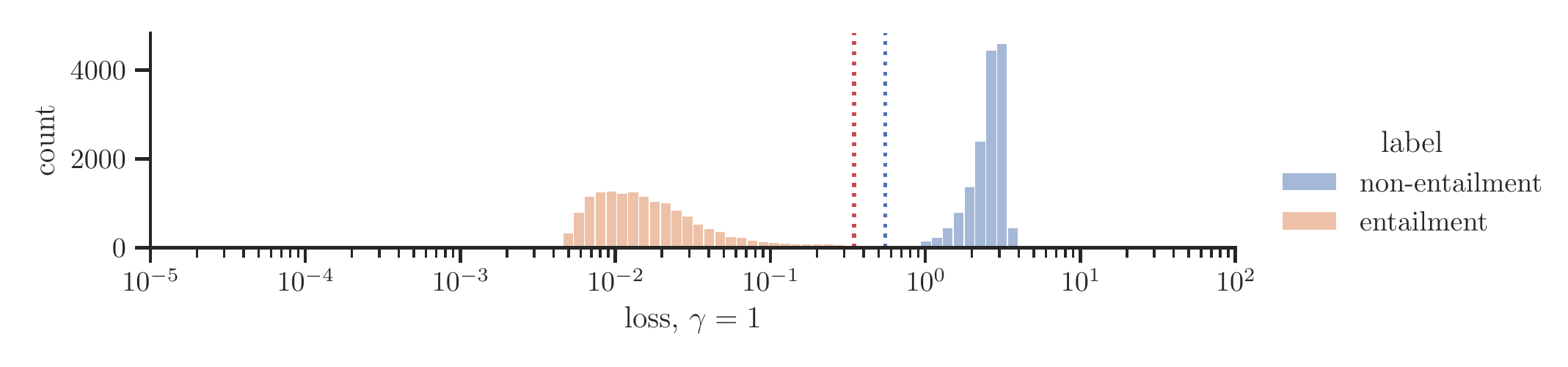}
    \includegraphics[width=0.75\columnwidth]{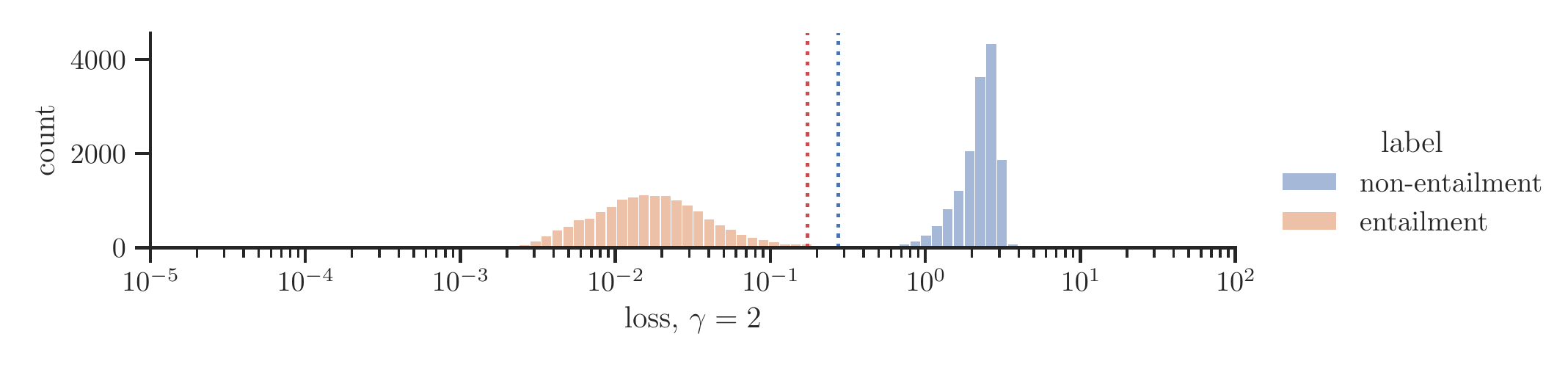}
    \includegraphics[width=0.75\columnwidth]{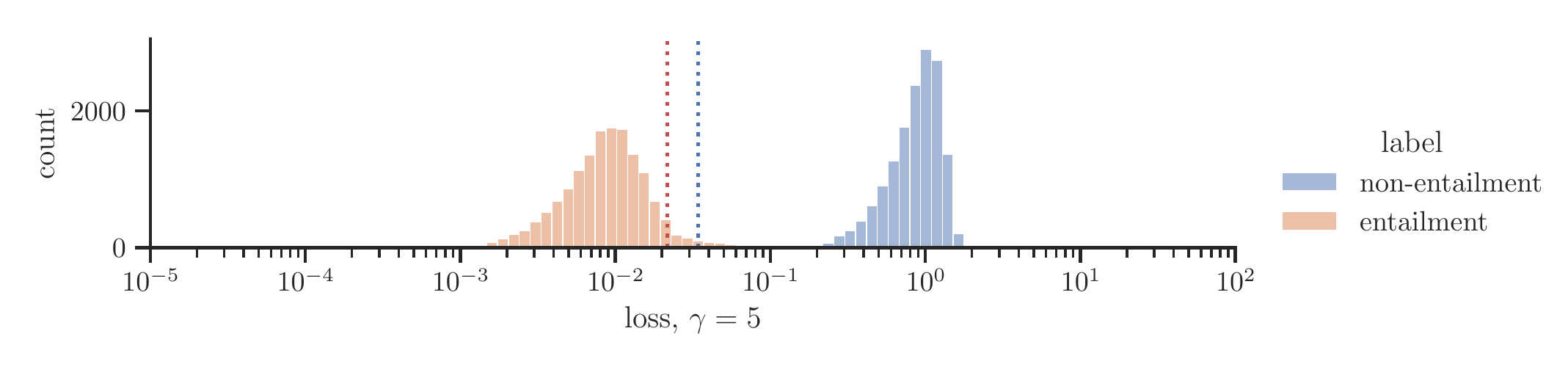}
    \includegraphics[width=0.75\columnwidth]{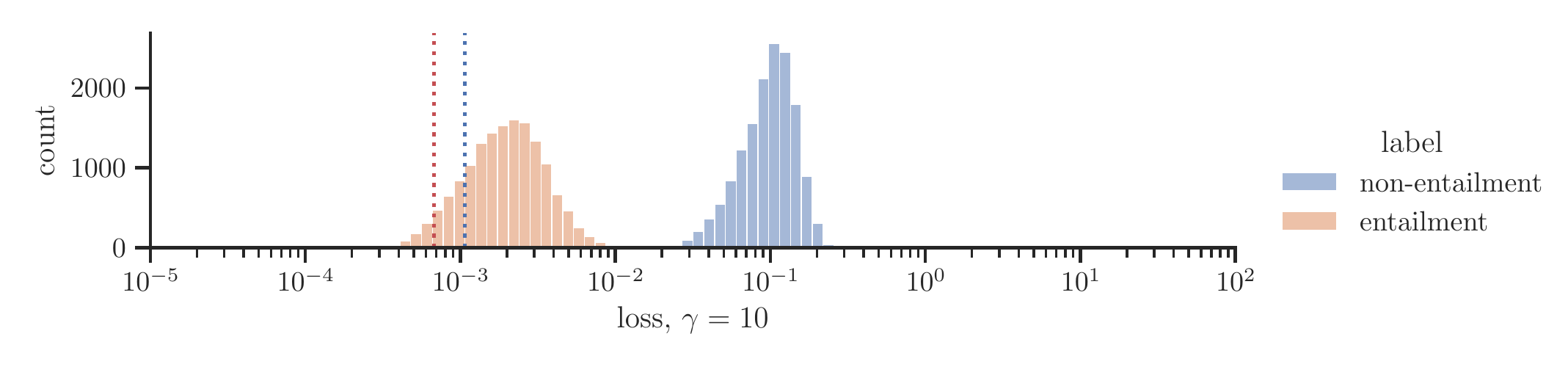}
    \caption{
    The distribution of loss of the \ac{HANS} test set produced by the BERT model for increasing values of $\gamma$. The blue line designates the loss for $0.33$ probability given to the ground truth class, which is the lowest possible probability for correct classification with three-label problems. Similarly, the red line designates the $0.5$ correct class prediction probability, the probability at which correct classification is guaranteed.
    }
    \label{fig:plot-a-hans-valid-v1}
\end{figure}

\begin{figure}
    \centering
    \includegraphics[width=\columnwidth]{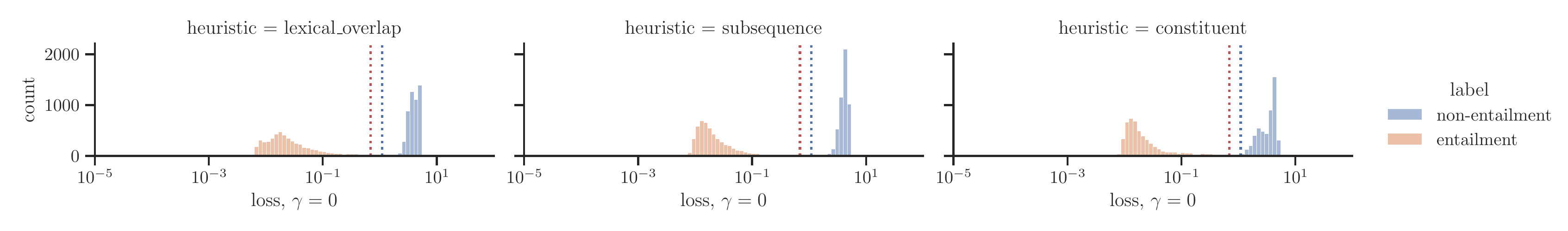}
    \includegraphics[width=\columnwidth]{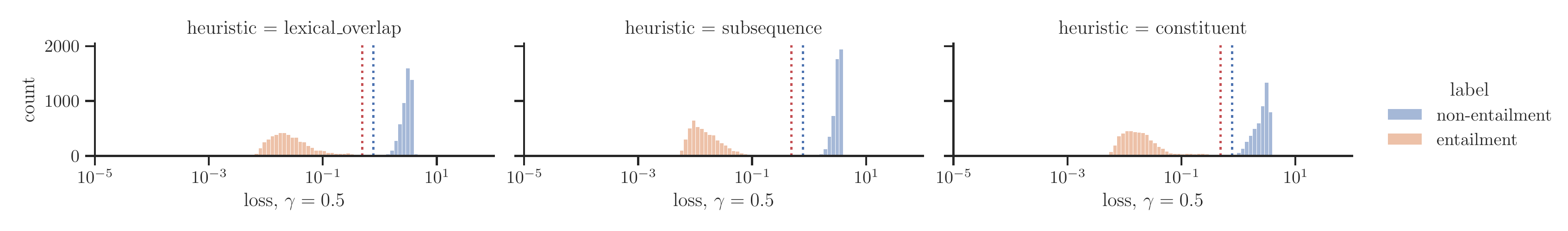}
    \includegraphics[width=\columnwidth]{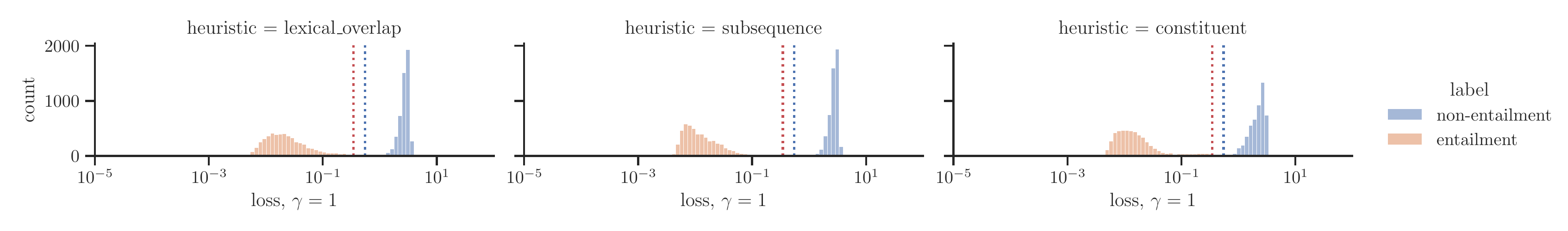}
    \includegraphics[width=\columnwidth]{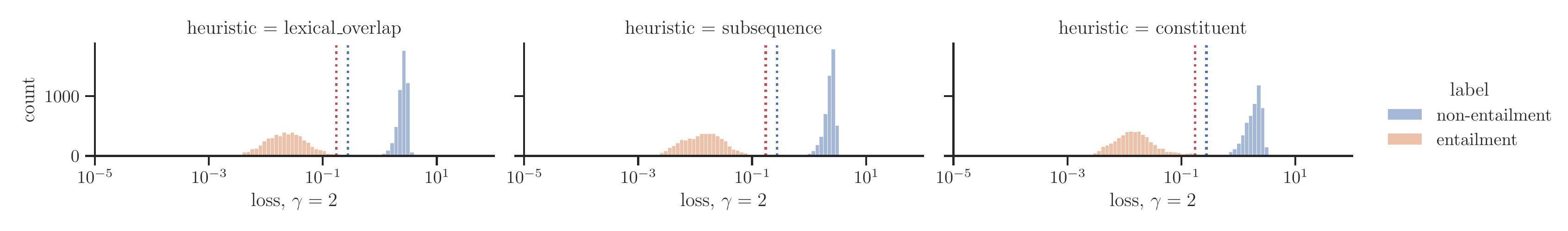}
    \includegraphics[width=\columnwidth]{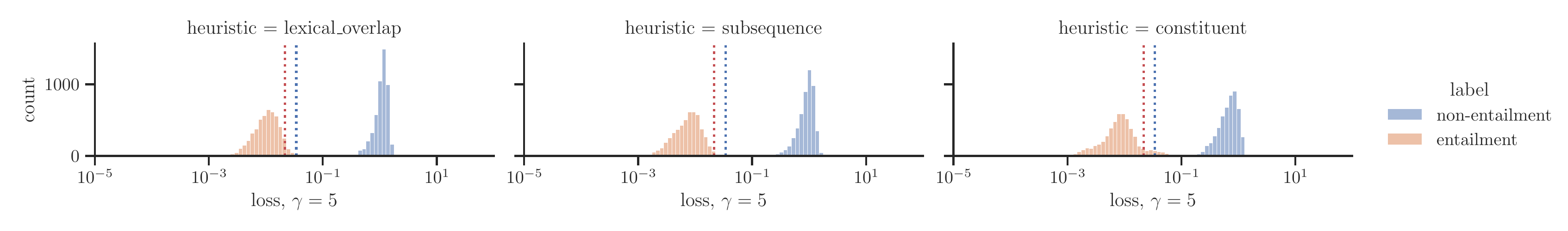}
    \includegraphics[width=\columnwidth]{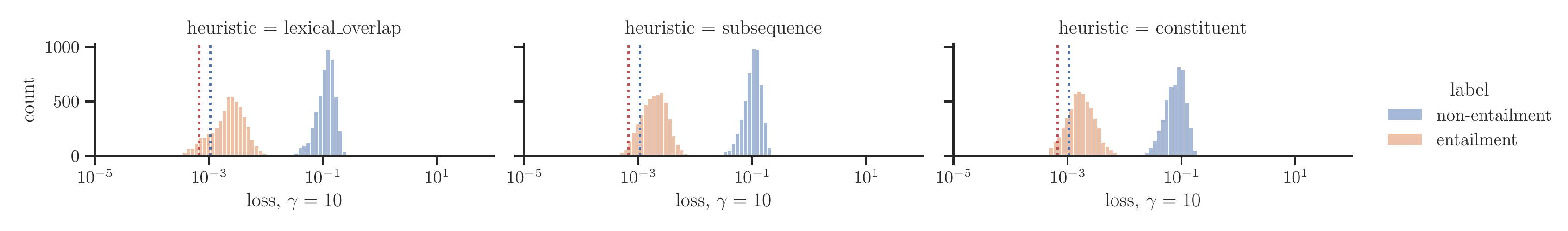}
    \caption{
    The distribution of loss of the \ac{HANS} set per heuristic produced by the BERT model for increasing values of $\gamma$. The blue line designates the loss for $0.33$ probability given to the ground truth class, which is the lowest possible probability for correct classification with three-label problems. Similarly, the red line designates the $0.5$ correct class prediction probability, the probability at which correct classification is guaranteed.
    }
    \label{fig:plot-a-hans-valid-v2}
\end{figure}

\begin{figure}
    \centering
    \includegraphics[width=0.75\columnwidth]{imgs_short/154.pdf}
    \includegraphics[width=0.75\columnwidth]{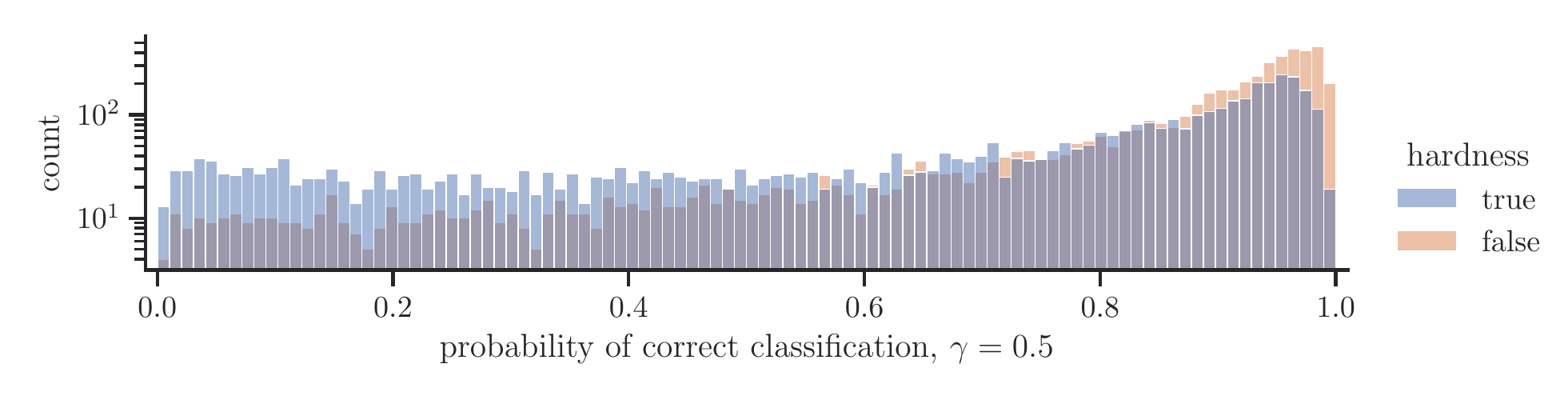}
    \includegraphics[width=0.75\columnwidth]{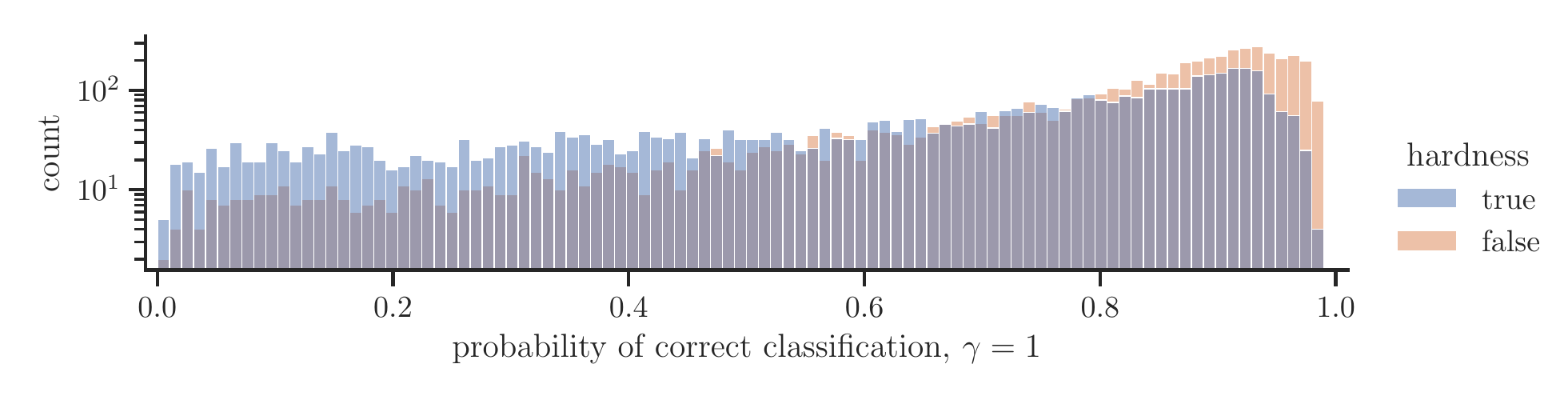}
    \includegraphics[width=0.75\columnwidth]{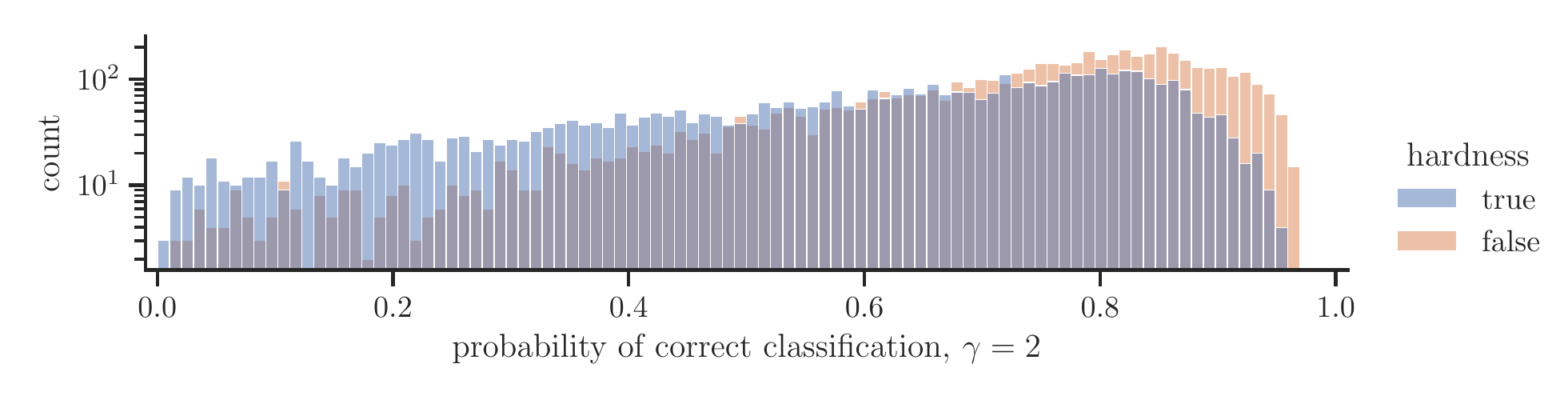}
    \includegraphics[width=0.75\columnwidth]{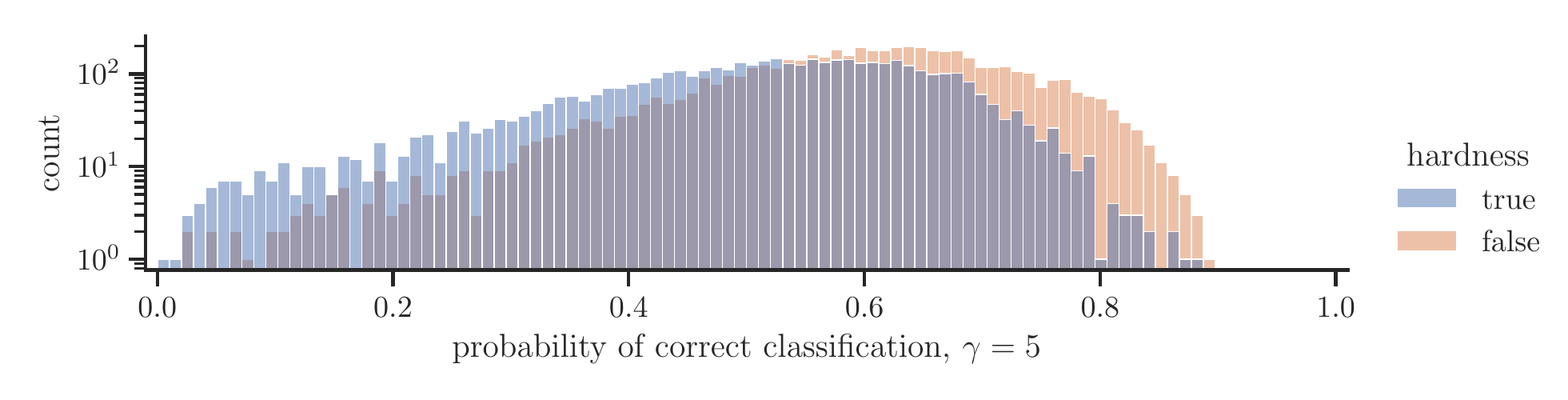}
    \includegraphics[width=0.75\columnwidth]{imgs_short/158.pdf}
    \caption{The distribution of ground truth class probabilities of the \ac{MNLI} test set produced by the BERT model for increasing values of $\gamma$. Hardness of samples is defined following \cite{gururangan2018annotation}. The tendency of the probabilities to converge close to $0.5$ when increasing gamma is a direct consequence of the focal loss's design, and can be a positive effect for avoiding learning heuristics by focusing on harder examples, but also removes certainty from correctly classified data.
    }
    \label{fig:plot-b-mnli-valid-v1}
\end{figure}

\begin{figure}
    \centering
    \includegraphics[width=\columnwidth]{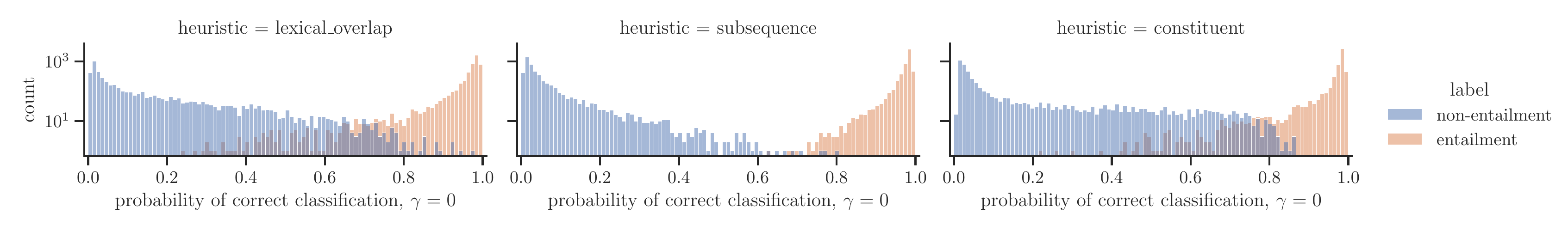}
    \includegraphics[width=\columnwidth]{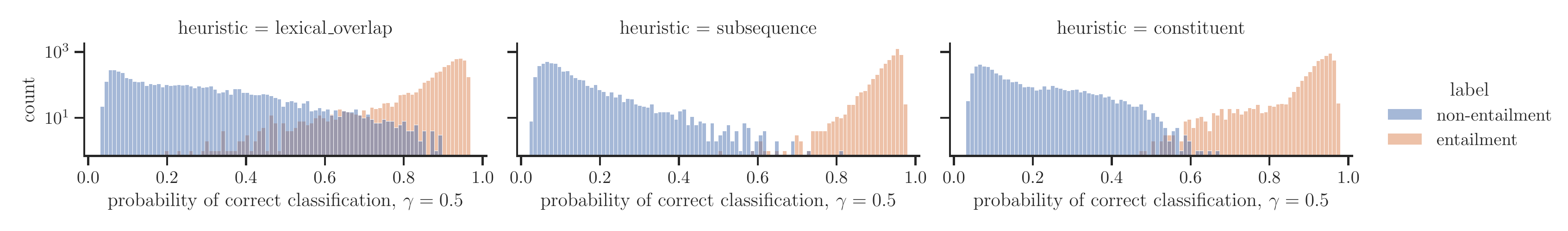}
    \includegraphics[width=\columnwidth]{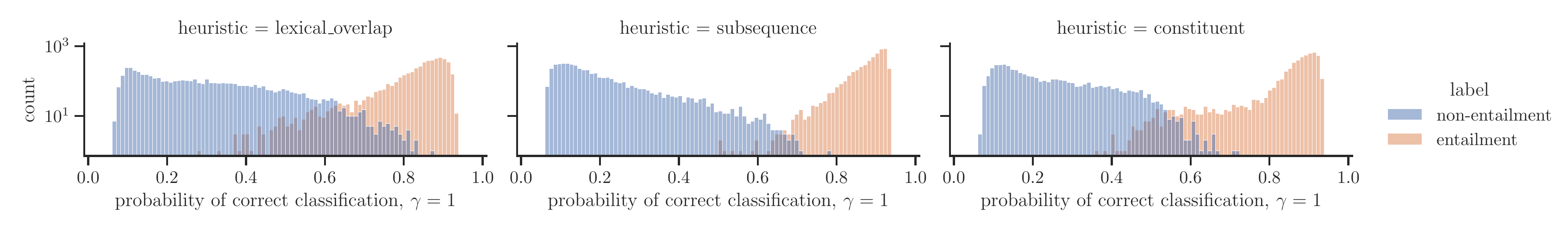}
    \includegraphics[width=\columnwidth]{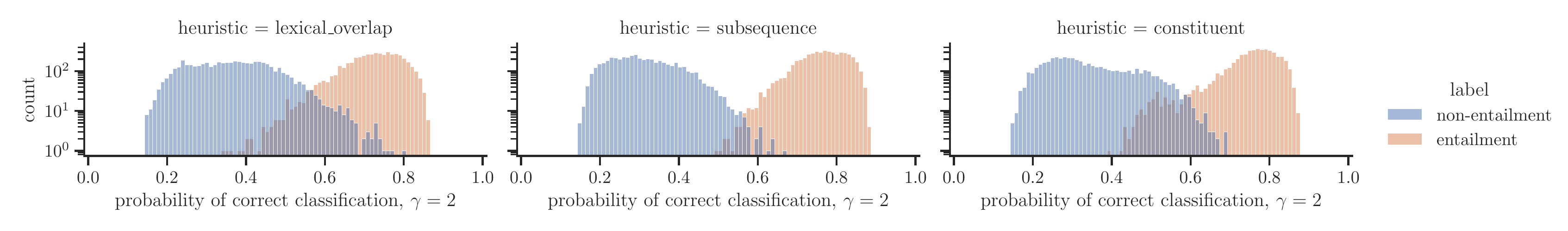}
    \includegraphics[width=\columnwidth]{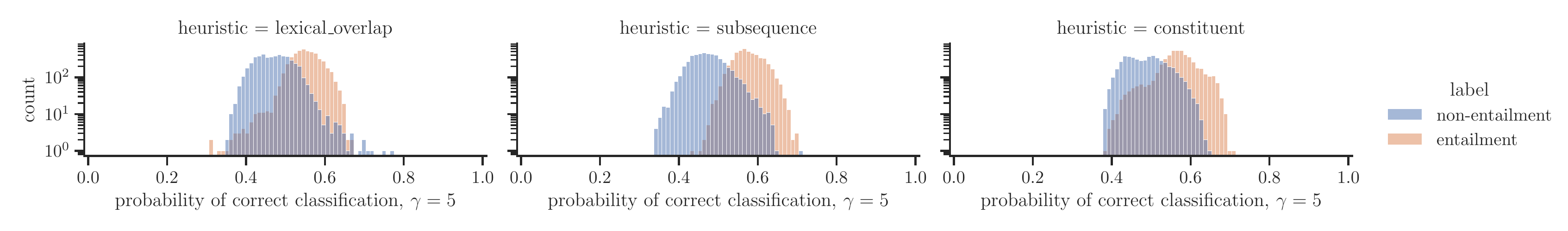}
    \includegraphics[width=\columnwidth]{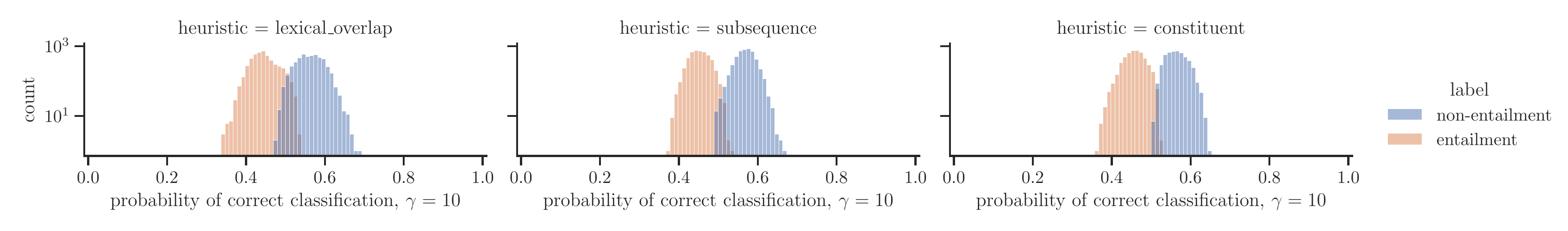}
    \caption{
    The distribution of ground truth class probabilities on the \ac{HANS} test per heuristic produced by the BERT model for increasing values of $\gamma$. The tendency of the probabilities to converge close to $0.5$ when increasing gamma is a direct consequence of the focal loss's design, and can be a positive effect for avoiding learning heuristics by focusing on harder examples, but also removes certainty from correctly classified data.
    }
    \label{fig:plot-b-hans-valid-v2}
\end{figure}

The distribution of loss for the on the MNLI train dataset is shown in Figure \ref{fig:plot-a-mnli-train-v1}, on MNLI test in Figure \ref{fig:plot-a-mnli-valid-v1} and Figure \ref{fig:plot-a-mnli-valid-v2}, and on HANS in Figure \ref{fig:plot-a-hans-valid-v1} and Figure \ref{fig:plot-a-hans-valid-v2}. The distribution of ground truth class probabilities on the MNLI test set is shown in Figure \ref{fig:plot-b-mnli-valid-v1}, and on HANS in Figure \ref{fig:plot-b-hans-valid-v1} and \ref{fig:plot-b-hans-valid-v2}.


\begin{figure}
    \centering
    \includegraphics[width=0.75\columnwidth]{imgs_short/210.pdf}
    \includegraphics[width=0.75\columnwidth]{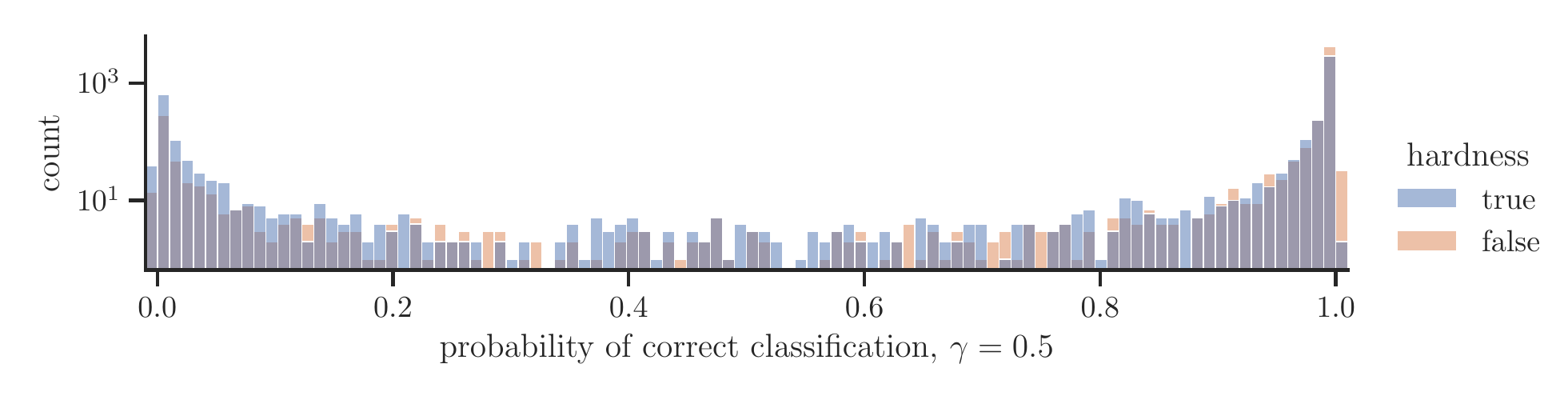}
    \includegraphics[width=0.75\columnwidth]{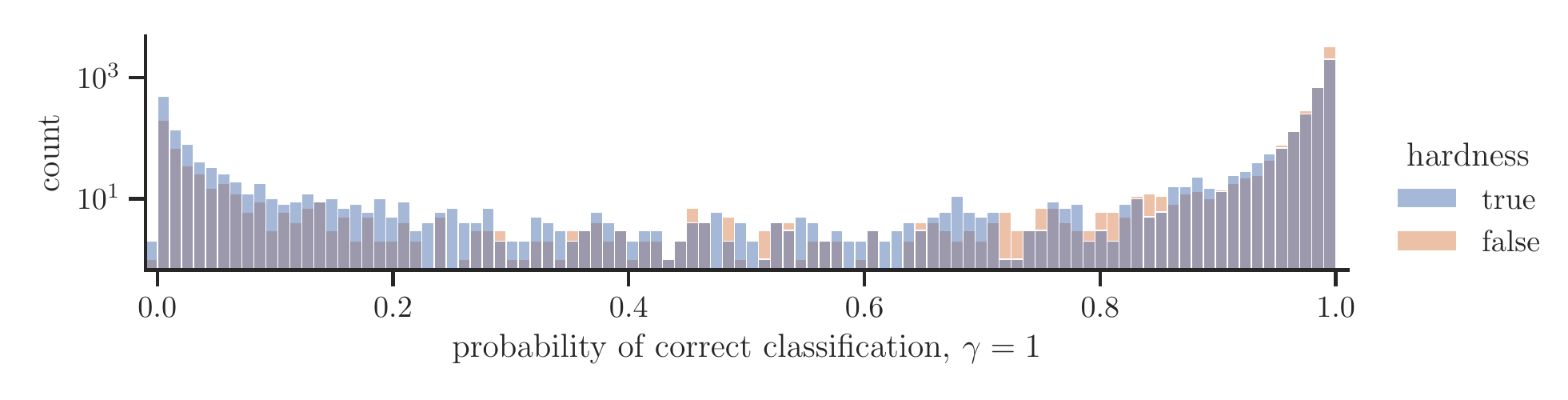}
    \includegraphics[width=0.75\columnwidth]{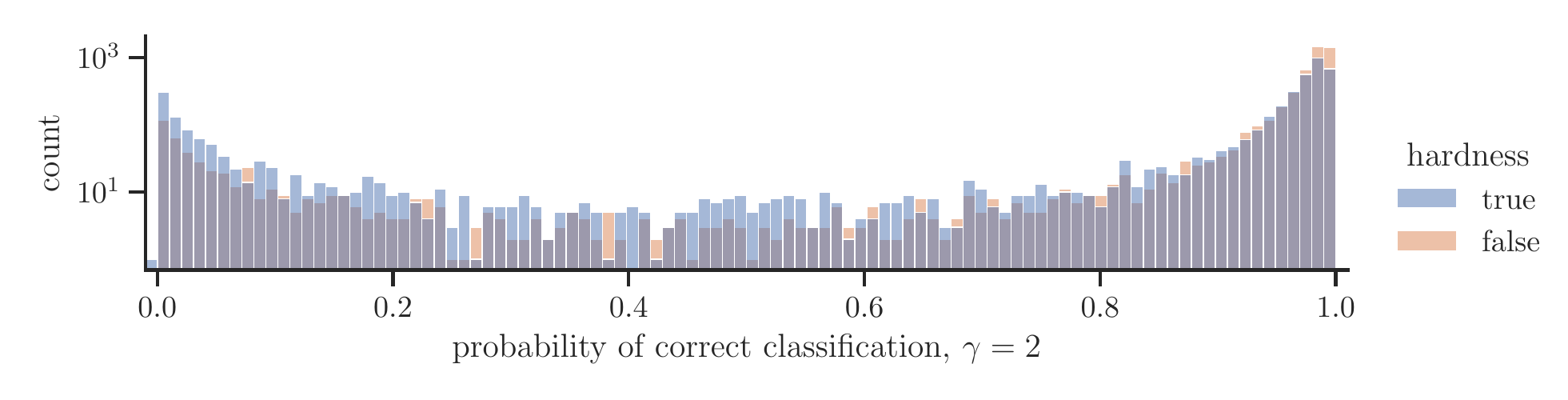}
    \includegraphics[width=0.75\columnwidth]{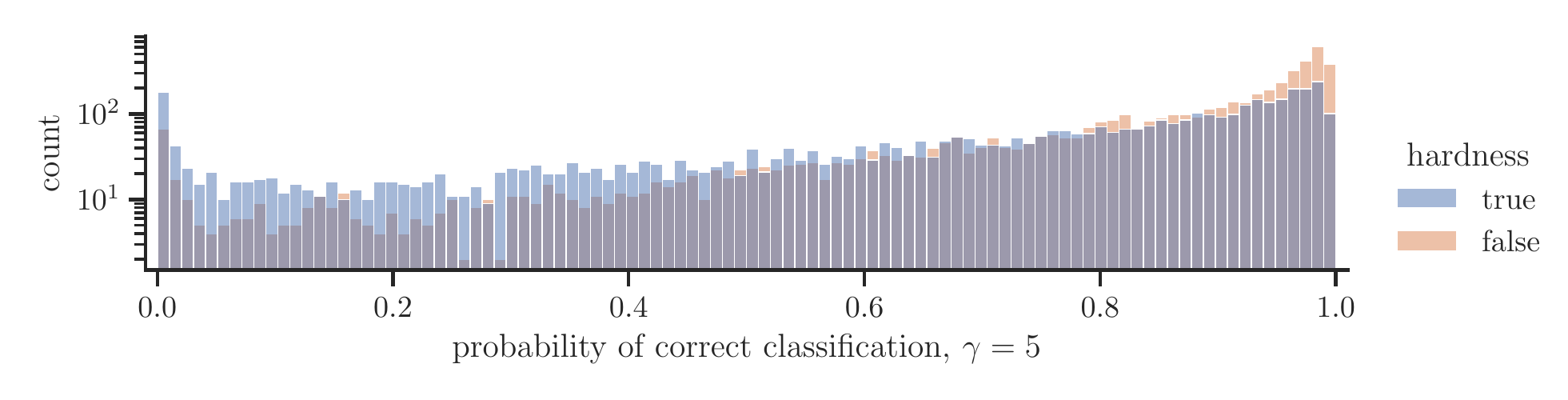}
    \includegraphics[width=0.75\columnwidth]{imgs_short/212.pdf}
    \caption{
    The distribution of ground truth class probabilities on the \ac{MNLI} test set produced by the BERT model when \textbf{no early stopping} was performed, for increasing values of $\gamma$. Hardness of samples is defined following \cite{gururangan2018annotation}. The tendency of the probabilities to converge close to $0.5$ when increasing gamma is a direct consequence of the focal loss's design, and can be a positive effect for avoiding learning heuristics by focusing on harder examples, but also removes certainty from correctly classified data.
}

    \label{fig:plot-b-hans-valid-v1}
\end{figure}
\begin{figure}
    \centering
    \includegraphics[width=0.75\columnwidth]{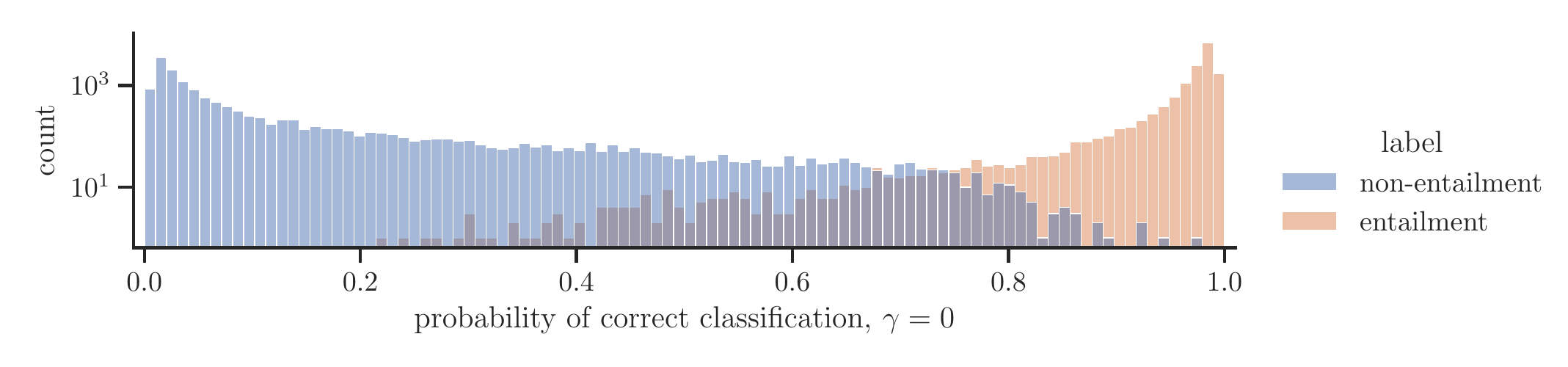}
    \includegraphics[width=0.75\columnwidth]{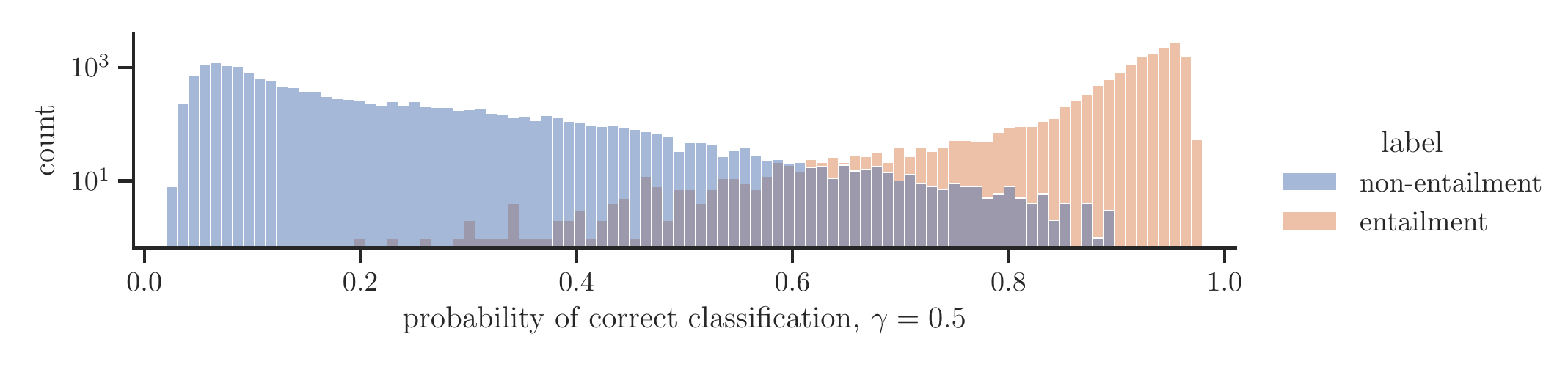}
    \includegraphics[width=0.75\columnwidth]{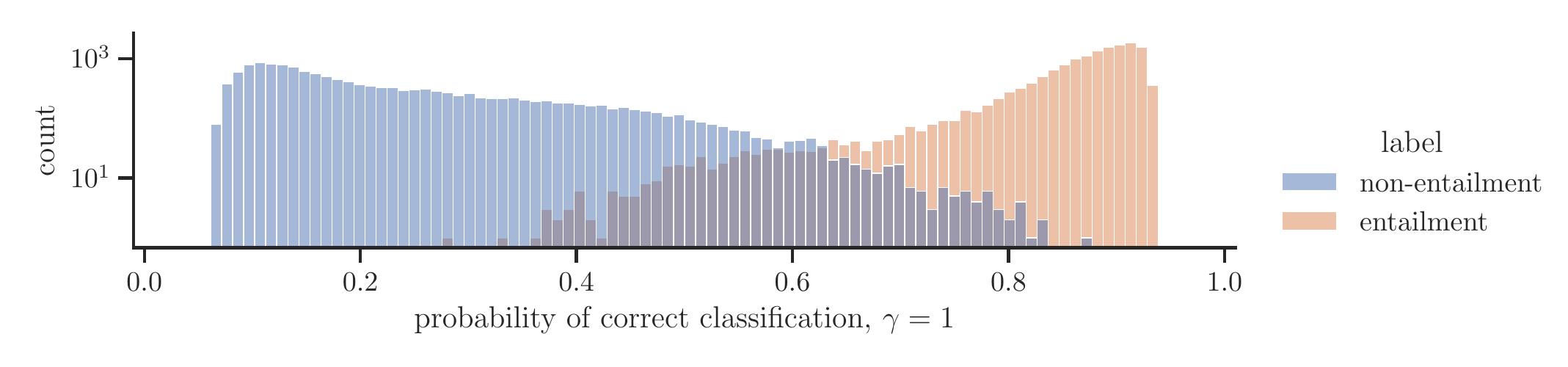}
    \includegraphics[width=0.75\columnwidth]{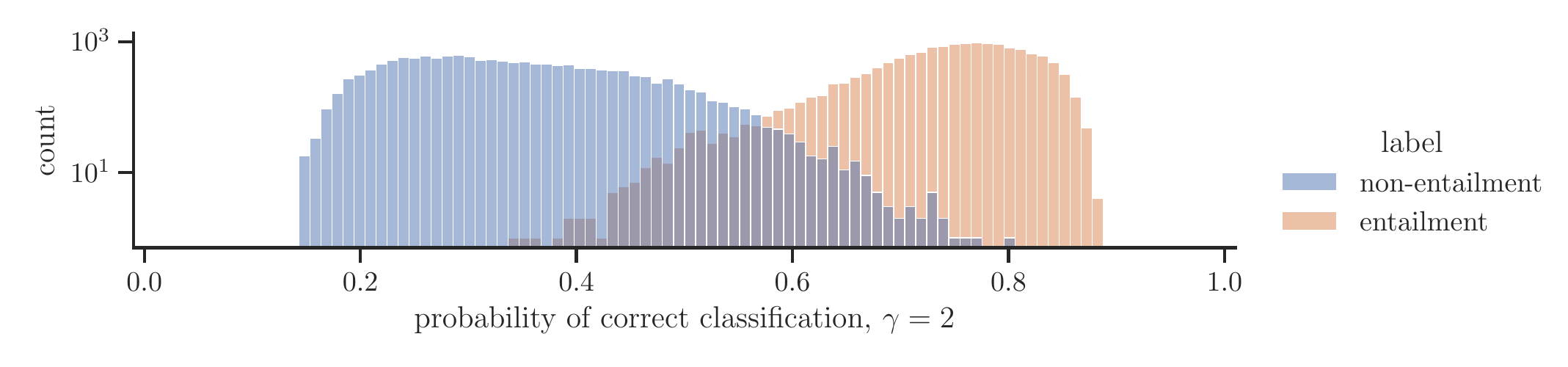}
    \includegraphics[width=0.75\columnwidth]{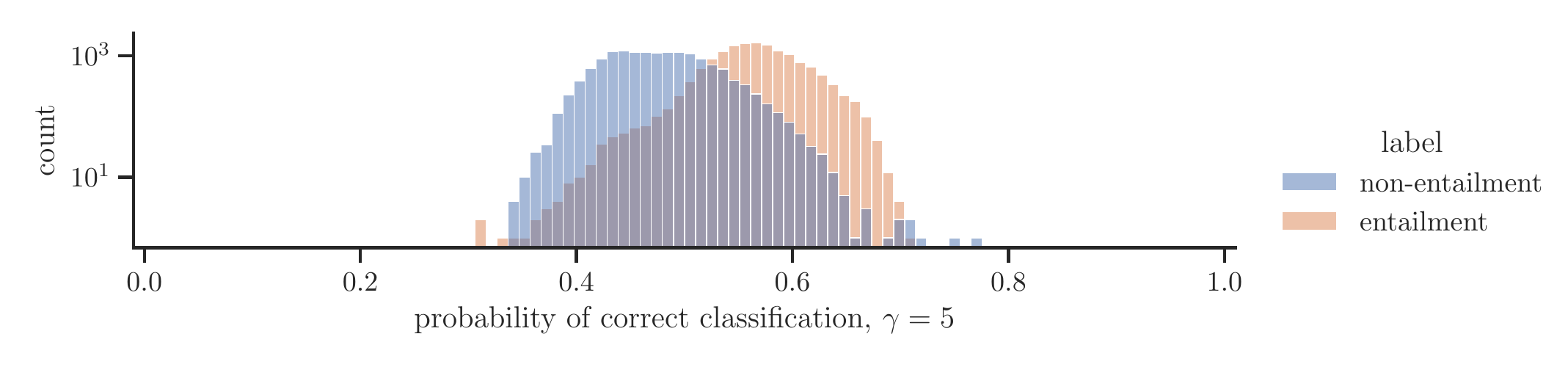}
    \includegraphics[width=0.75\columnwidth]{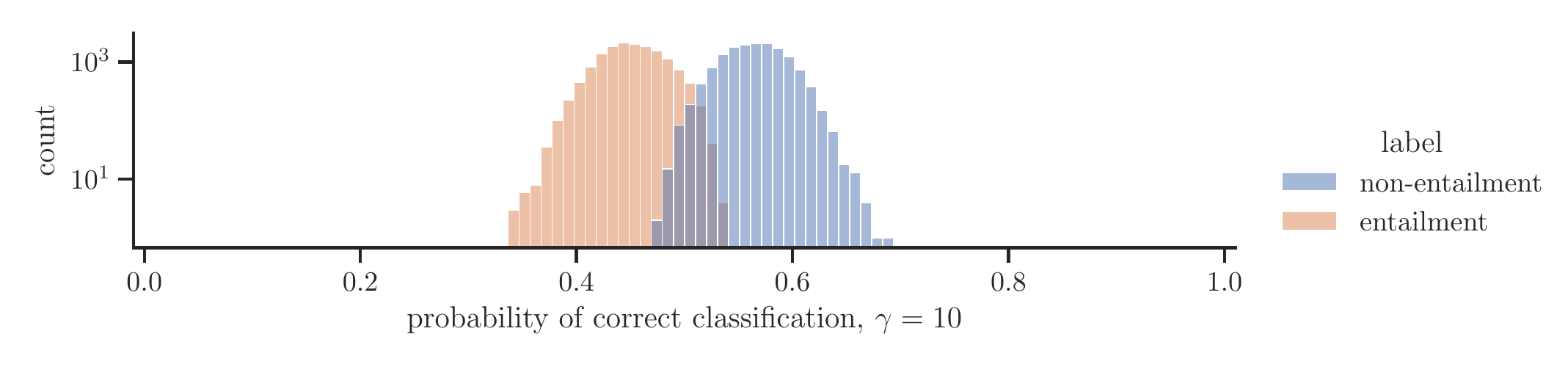}
    \caption{
    The distribution of ground truth class probabilities on the \ac{HANS} test produced by the BERT model for increasing values of $\gamma$. The tendency of the probabilities to converge close to $0.5$ when increasing gamma is a direct consequence of the focal loss's design, and can be a positive effect for avoiding learning heuristics by focusing on harder examples, but also removes certainty from correctly classified data.
}

    \label{fig:plot-b-hans-valid-v1}
\end{figure}

\end{document}